%% file: main.tex
\def\year{2022}\relax
\documentclass[letterpaper]{article} % DO NOT CHANGE THIS
\usepackage{aaai22}  % DO NOT CHANGE THIS
\usepackage{times}  % DO NOT CHANGE THIS
\usepackage{helvet}  % DO NOT CHANGE THIS
\usepackage{courier}  % DO NOT CHANGE THIS
\usepackage[hyphens]{url}  % DO NOT CHANGE THIS
\usepackage{graphicx} % DO NOT CHANGE THIS
\usepackage{natbib}  % DO NOT CHANGE THIS AND DO NOT ADD ANY OPTIONS TO IT
\usepackage{caption} % DO NOT CHANGE THIS AND DO NOT ADD ANY OPTIONS TO IT
\DeclareCaptionStyle{ruled}{labelfont=normalfont,labelsep=colon,strut=off} % DO NOT CHANGE THIS
\usepackage{algorithm}
\usepackage{algorithmic}

\usepackage{epsfig}
\usepackage{graphicx}
\usepackage{amsmath}
\usepackage{amssymb}
\usepackage{subcaption}
\usepackage{booktabs}
\usepackage{adjustbox}
\usepackage{enumitem}
\usepackage{multirow}
\usepackage[usenames,dvipsnames,svgnames,table]{xcolor}
\makeatletter
\def\mathcolor#1#{\@mathcolor{#1}}
\def\@mathcolor#1#2#3{%
  \protect\leavevmode
  \begingroup
    \color#1{#2}#3%
  \endgroup
}  
\makeatother  
\usepackage{tablefootnote}
\usepackage{hyperref}
\usepackage{newfloat}
\usepackage{listings}
\floatstyle{ruled}
\newfloat{listing}{tb}{lst}{}
\floatname{listing}{Listing}
\title{Imaginative Walks: Generative Random Walk Deviation Loss for Improved  Unseen Learning Representation}
\author{
    Divyansh Jha\footnote{denotes equal contributions.}, Kai Yi$^*$, Ivan Skorokhodov, Mohamed Elhoseiny\\
}

\affiliations {
    King Abdullah University of Science and Technology (KAUST)\\
    \{divyansh.jha, kai.yi, ivan.skorokhodov, mohamed.elhoseiny\}@kaust.edu.sa\\
}

\usepackage{bibentry}
\begin{document}

\maketitle
   
\begin{abstract}
% We propose a novel loss for generative models, dubbed as GRaWD (Generative Random Walk  Deviation), to improve learning representations of unexplored visual spaces.  Quality learning representation of unseen classes (or styles) is critical to facilitate novel image generation and better generative understanding of unseen visual classes (a.k.a. Zero-Shot Learning, ZSL). By generating representations of unseen classes from their semantic descriptions, such as attributes or text, Generative ZSL aims at identifying unseen categories discriminatively from seen ones.    
% We define GRaWD by constructing a dynamic graph, including the seen class/style centers and generated samples in the current mini-batch.  Our loss starts a random walk probability from each center through visual generations produced from hallucinated unseen classes. As a deviation signal, we encourage the random walk to eventually land after $t$ steps in a feature representation that is hard to classify to any of the seen classes. We show that our loss can improve unseen class representation quality on four text-based  ZSL benchmarks on CUB and NABirds datasets and three attribute-based ZSL benchmarks on AWA2, SUN, and aPY datasets. We also study our loss's ability to produce meaningful novel visual art generations on the WikiArt dataset. Our experiments and human studies show that our loss can improve StyleGAN1 and StyleGAN2  generation quality, creating novel art that is significantly more preferred.

{We propose a novel loss for generative models, dubbed as GRaWD (Generative Random Walk  Deviation), to improve learning representations of unexplored visual spaces. Quality learning representation of unseen classes (or styles) is critical to facilitate novel image generation and better generative understanding of unseen visual classes, i.e., zero-shot learning (ZSL). By generating representations of unseen classes based on their semantic descriptions, e.g., attributes or text, generative ZSL attempts to differentiate unseen from seen categories.
The proposed GRaWD loss is defined by constructing a dynamic graph that includes the seen class/style centers and generated samples in the current minibatch. Our loss initiates a random walk probability from each center through visual generations produced from hallucinated unseen classes. As a deviation signal, we encourage the random walk to eventually land after $t$ steps in a feature representation that is difficult to classify as any of the seen classes. We demonstrate that the proposed loss can improve unseen class representation quality inductively on text-based ZSL benchmarks on CUB and NABirds datasets and attribute-based ZSL benchmarks on AWA2, SUN, and aPY datasets. In addition, we investigate the ability of the proposed loss to generate meaningful novel visual art on the WikiArt dataset. The results of experiments and human evaluations demonstrate that the proposed GRaWD loss can improve StyleGAN1 and StyleGAN2 generation quality and create novel art that is significantly more preferable.  % Our code will be made publicly available. 
Our code is made publicly available at \url{https://github.com/Vision-CAIR/GRaWD}.

% \href{https://github.com/Vision-CAIR/GRaWD}{https://github.com/Vision-CAIR/GRaWD}.
}
\end{abstract}

%%%%%%%%% BODY TEXT
\section{Introduction}

\begin{figure}[t!]
 \centering
     \vspace{-2.5mm}
    \includegraphics[width=7.5cm]{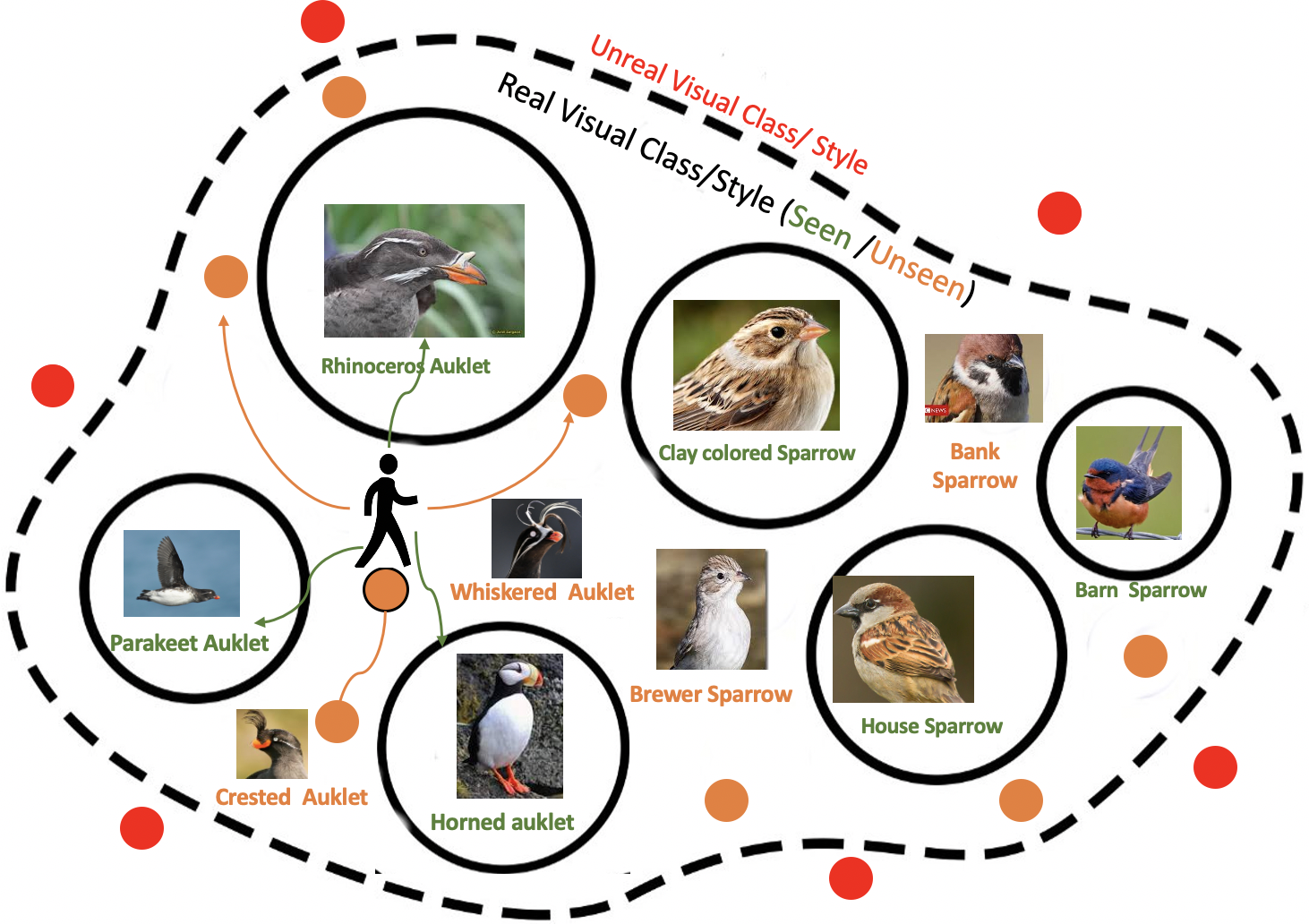}
    \vspace{-2mm}
    \caption{GRaWD loss encourages generatively visiting the \textcolor{orange}{orange} realistic space,  aiming to deviate from the seen classes and avoid the less real \textcolor[rgb]{1,0,0}{red} space. Our loss starts  from each seen class ( in   \textcolor[rgb]{0,0.7,0}{green}) and performs a random walk through generated examples of   hallucinated unseen classes (in \textcolor{orange}{orange}) for $T$ steps.  We then  encourage the landing representation to be  distinguishable from seen classes. With this property, the proposed loss helps improve \emph{generalized ZSL} performance and novel art creation. } 
           \label{fig1a}
               \vspace{-1mm}
           \vspace{-2.0ex}
\end{figure}

Generative models like GANs~\cite{goodfellow2014generative} and VAEs~\cite{kingma2013auto} are excellent tools for generating realistic images due to their ability to represent high-dimensional probability distributions. However, they are not explicitly trained to go beyond distribution seen during training. In recent years, generative models have been adopted to go beyond training data distributions and improve unseen class recognition (also known as zero-shot learning)~\cite{guo2017synthesizing,long2017zero, guo2017zero,kumar2018generalized,Elhoseiny_2018_CVPR,vyas2020leveraging}. These methods train a conditional generative model $G(s_k, z )$~\cite{mirza2014conditional,odena2016conditional}, where    $s_k$ is the semantic description of class $k$ (attributes or text descriptions) and  $z$ represents within-class variation (e.g., $z \in \mathcal{N}(0, I)$). After training, $G(s_k,z)$ is used to generate imaginary data for unseen classes transforming ZSL into a traditional classification task trained on the generated data. Understanding unseen classes is mainly leveraged by the generative model's improved ability to produce discriminative visual features/representations using $G(s_u,z)$ from their corresponding unseen semantic descriptions (i.e., $s_u$).
\begin{figure}[t!]
 \centering
    \includegraphics[width=9.0cm, trim=120 0 120 0, clip]{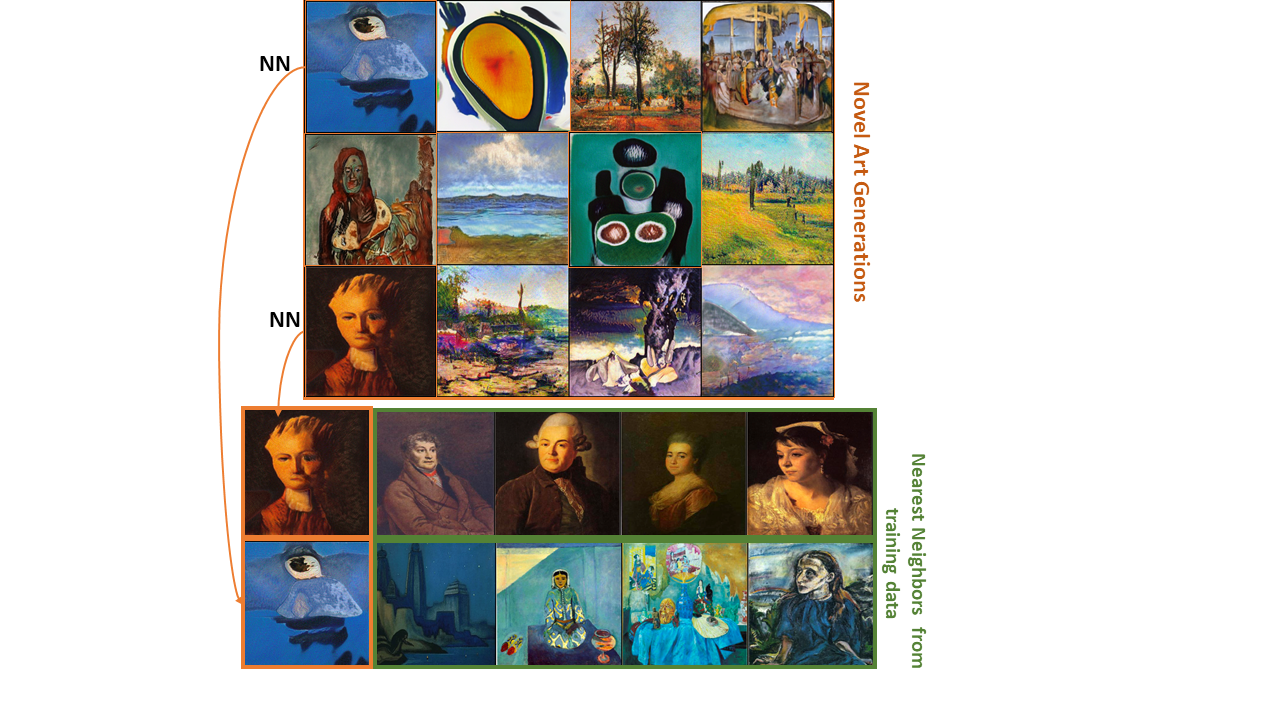}
    \vspace{-7ex}
    \caption{Art images on top with  \textcolor{orange}{orange} borders are generated using our loss. The bottom part  shows the Nearest Neighbors (NN) in the training set (with \textcolor[rgb]{0,0.7,0}{green} borders), which are  different indicating our generations' novelty.  }
    \label{fig1b}
    \vspace{-2.5ex}
\end{figure}

To generate likable novel visual content, GANs' training has been augmented with a loss that encourages careful deviation from existing classes~\cite{can_2017,sbai2018design,hertzmann2018can}. Such models were shown to have some capability to produce unseen aesthetic art~\cite{can_2017}, fashion~\cite{sbai2018design, wu2021clothgan}, and design~\cite{nobari2021creativegan}. In a generalized ZSL context,   CIZSL~\cite{elhoseiny2019creativity}  showed an improved performance by modeling a similar deviation to encourage discrimination explicitly between seen and unseen classes. %The main idea of these losses is to improve unseen representation quality by encouraging the produced visual generations to be distinguishable from seen classes in ZSL in~\cite{elhoseiny2019creativity},  seen/known painting styles in art generation~\cite{can_2017}, and known texture and shape categories for  fashion~\cite{sbai2018design}. 
These losses improve unseen representation quality by encouraging the produced visual generations to be distinguishable from seen classes in ZSL~\cite{elhoseiny2019creativity} and seen styles in art ~\cite{can_2017} and fashion~\cite{sbai2018design} generation.

  We propose Generative Random Walk Deviation Loss (\emph{GRaWD}) as a \emph{parameter-free} graph-based loss to improve learning representation of unseen classes; see Fig.~\ref{fig1a}. Our loss starts from each seen class (in green) and performs a random walk through the generated examples of hallucinated unseen classes (in orange) for $T$ steps. Then, we encourage the landing representation to be distant and distinguishable from the seen class centers. GRaWD loss is computed over a similarity graph involving seen class centers and generated examples in the current minibatch of hallucinated unseen classes. Thus, GRaWD takes a global view of the data manifold compared to existing deviation losses that are local/per example (e.g., ~\citet{sbai2018design,can_2017,elhoseiny2019creativity}). In contrast to transductive methods (e.g., ~\citet{vyas2020leveraging}),  our loss is purely inductive; therefore, does not require real descriptions of unseen classes during training. Our work is connected to recent advances in semi-supervised learning (e.g.,~\citet{zhang2018metagan,ayyad2020semi,ren2018metalearning,8099557,li2019learning}) that leverage unlabeled data within the training classes. In these methods, unlabeled data are encouraged to be attracted to existing classes. Our goal is the \emph{opposite}, deviating from seen classes. Also, our loss operates on generated data of hallucinated unseen classes instead of provided unlabeled data.

\noindent \textbf{Contribution.} 
We propose a generative random walk loss that leverages generated data by exploring the unseen embedding space discriminatively against the seen classes; see Fig. ~\ref{fig1a}.  Our loss is unsupervised on the generative space and can be applied to any GAN architecture (e.g., DCGAN~\cite{radford2015unsupervised}, StyleGAN  ~\cite{karras2019style}, and StyleGAN2 ~\cite{karras2020analyzing}).
We show that our  GRaWD loss helps understand unseen visual classes better, improving generalized zero-shot learning tasks on challenging benchmarks. We also show that compared to existing deviation losses, GRaWD improves the generative capability in unseen space of liked art; see Fig.~\ref{fig1b}.

% \kai{also in the introduction and method part, we need to emphasize the saying inductive zero-shot learning. may from the published tist-19 survey paper. Whether classifier-based model should be regarded as trasductive in our paper.}

\section{Related Work}

 \begin{figure*}[t!]
 \centering
          	\vspace{-3mm}
         	\includegraphics[width=0.78\textwidth]{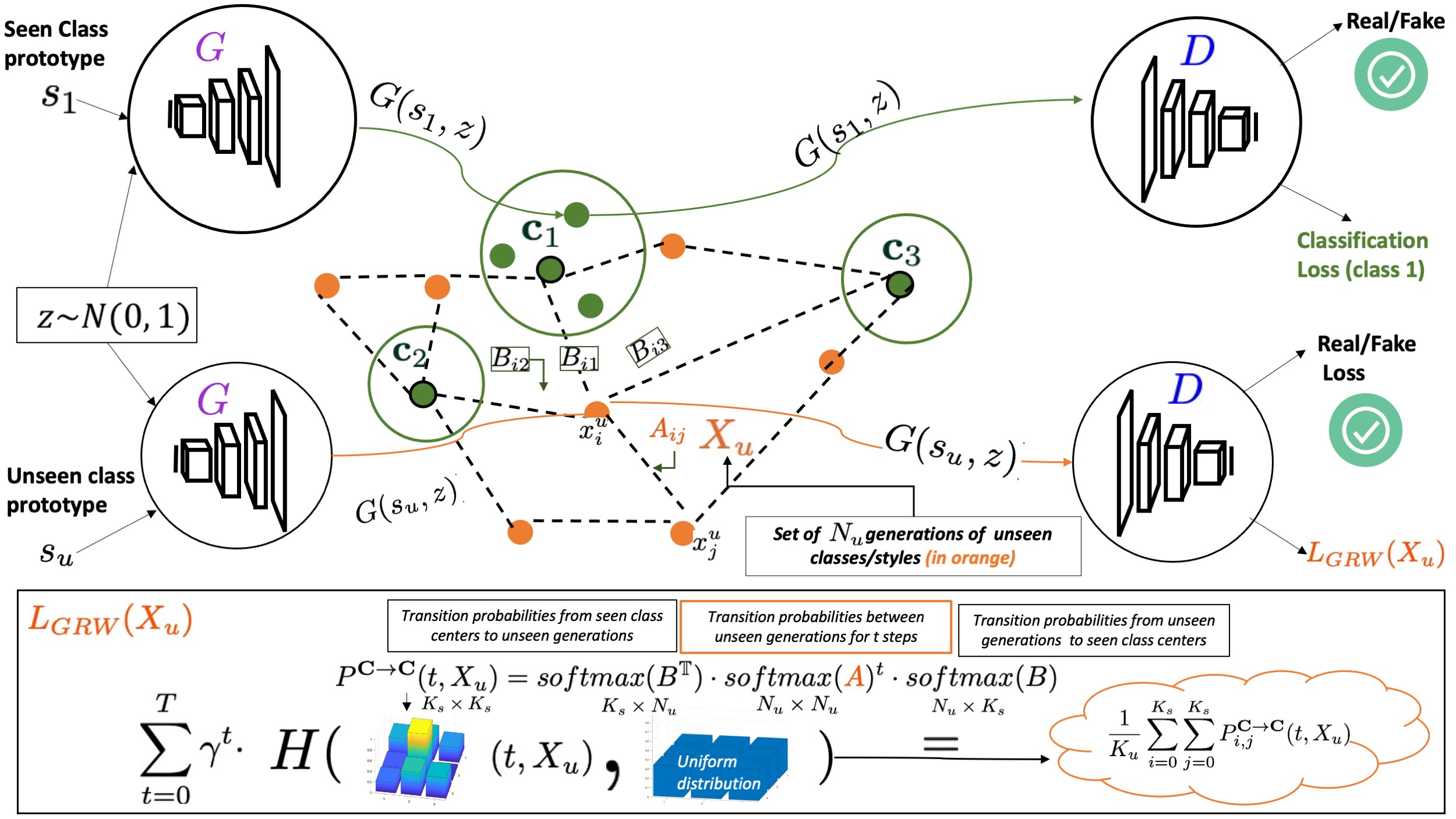}
         	\vspace{-2mm}
         \caption{Generative Random Walk Deviation loss starts from each seen class center (i.e., $\mathbf{c}_i$). It then performs a random walk through generated examples of hallucinated unseen classes using $G(s_u,z)$ for $T$ steps. The landing probability distribution of the random walk is encouraged to be uniform over the seen classes. For careful deviation from seen classes,  the generated images are encouraged to be classified as real by the Discriminator $D$; see Eq.~\ref{eq:RW_loss}. } 
           \label{fig1approach}
                   	\vspace{-2mm}
\end{figure*}

\textbf{Generative Models with Deviation Losses. }
In the context of computational creativity, several approaches have been proposed to produce original items with aesthetic and meaningful characteristics ~\cite{machado2000nevar,mordvintsev2015inceptionism,dipaola2009incorporating,tendulkar2019trick}.  Various early studies have made progress on writing pop songs~\cite{briot2017deep}, and transferring  styles of great painters  to other images ~\cite{Gatys2016ImageStyleTransfer,Date2017Fashioning,dumoulin2016learned,johnson2016perceptual,Isola2016ImageToImage} or doodling sketches~\cite{ha2017neural}. The creative space of the style transfer images is limited by the content image and the transfer image, which could be an artistic image by Van Gogh.  GANs~\cite{goodfellow2014generative,radford2015unsupervised,ha2017neural,Reeds2016whatwhere,zhang2016stackgan,karras2017progressive,karras2019style} have a capability to learn visual distributions and produce images from a latent $z$ vector. However,  they are not trained explicitly to produce novel content beyond the training data. More recent work explored an early capability to produce novel art with CAN~\cite{elgammal2017can}, and fashion designs with a holistic CAN (an improved version of CAN)~\cite{sbai2018design}, which are based on augmenting DCGAN~\cite{radford2015unsupervised} with a  loss encouraging deviation from existing styles.  The difference between CAN and holistic-CAN is that the deviation signal is Binary Cross Entropy over individual styles for CAN~\cite{elgammal2017can} and  Multi-Class Cross Entropy (MCE) loss overall styles in Holistic-CAN~\cite{sbai2018design}.  Similar deviation losses were proposed in CIZSL~\cite{elhoseiny2019creativity} for ZSL.   

In contrast to these deviation losses, our loss is more global as it establishes dynamic messages between generations that are produced every mini-batch iteration and seen visual spaces. These generations should deviate from seen class spaces represented by class centers. In our experiments, we applied our loss on unseen class recognition and  producing novel visual generations, showing superior performance compared to existing losses. We also note that random walks have been explored in the literature in the context of semi-supervised and few-shot learning for attracting unlabeled data points to its corresponding class (e.g.~\citet{ayyad2020semi,8099557}).  In contrast, we develop a random walk-based method to deviate from seen classes, which is an opposite objective, and operates on  generated data rather than unlabeled data that are not available in purely inductive setups; see Fig.~\ref{fig1a}.      %ZSL

\noindent \textbf{Zero-Shot Learning Methods.}  Classical ZSL methods directly predict attribute confidence from images to facilitate zero-shot recognition  (e.g., seminal works by ~\citet{lampert2009, lampert2013attribute} and~\citet{farhadi2009describing}). Current ZSL methods can be classified into two branches. One branch casts the task as a visual-semantic embedding problem~\cite{frome2013devise,skorokhodov2021class,liu2020hyperbolic}. \citet{akata2015evaluation,akata2016label} proposed Attribute Label Embedding(ALE) to model visual-semantic embedding as a  bilinear compatibility function between the image space and the attribute space. In~\cite{zhang2016learning}, deep ZSL methods were presented to model the non-linear mapping between vision and class descriptions. In the context of ZSL from noisy textual descriptions, an early linear approach for Wikipedia-based ZSL was proposed in~\cite{elhoseiny2013write}. Orthogonal to these improvements, generative models like GANs~\cite{goodfellow2014generative} and VAEs~\cite{kingma2013auto} have been adopted to formulate multi-modality in zero-shot recognition by synthesizing visual features of unseen classes given its semantic description, e.g.,~\cite{kumar2018generalized,Elhoseiny_2018_CVPR, schonfeld2019generalized, narayan2020latent, han2021contrastive, chen2021free}. \citet{Elhoseiny_2018_CVPR} introduced a GAN model with a classification head with the standard real/fake head to improve text-based ZSL. \citet{schonfeld2019generalized} proposed a cross and distribution aligned VAE to better leverage the seen and unseen relationships. \citet{han2021contrastive} utilized a generative network along with a multi-level supervised contrastive embedding strategy to learn images and semantic relationships. Our GRaWD loss helps improve the out-of-distribution performance of generative ZSL  models.  %, elhoseiny2021cizsl++

\section{Approach}
\input{approach_new}
\begin{table*}[!htbp]
\parbox{.58\linewidth}{
\caption{Ablation studies on CUB Dataset (text).} 
\vspace{-1.5ex}
\centering
\resizebox{.58\textwidth}{!}{
\begin{tabular}{llccccc}\toprule
%&\multicolumn{4}{c}{\textbf{CUB Dataset - Ablation Study}} \\\cmidrule{2-5}
~ & \multirow{2}{*}{Setting} & \multicolumn{2}{c}{CUB-Easy} & \multicolumn{2}{c}{CUB-Hard} \\ \cmidrule{3-6}
~ & ~                        &Top-1 Acc (\%) & SU-AUC (\%) & Top1-Acc (\%) &SU-AUC (\%) \\\midrule

\multirow{2}{*}{Deviation losses} &+ GRaWT (T=0)               & 44.0 & 39.5 & 13.7 & 11.8 \\
~ & + GRaWT (T=3)                        & 43.4 & 38.8 & 13.2 & 11.4 \\ 
 on GAZSL~\cite{Elhoseiny_2018_CVPR} &  + Classify $G(s_u, z)$ as class $K^{s+1}$                        & 43.2 &  38.3& 11.31 & 9.5 \\ 
~ &+ CIZSL\cite{elhoseiny2019creativity}                       & 44.6 &  39.2 &  14.4  & 11.9 \\ 
\midrule
% \multirow{5}{*}{Walk length} & GRaWD & 45.4 & 15.5 & 40.7 & 13.7 \\ \midrule
\multirow{2}{*}{Walk length} &  + GRaWD (T=1) &45.41 &39.62 &13.79 &12.58 \\
~ &  + GRaWD (T=3) &45.11 &39.25 &14.21 &13.22 \\
 on GAZSL~\cite{Elhoseiny_2018_CVPR}  &  + GRaWD (T=5) &45.40 &40.51 &14.00 &13.07 \\
~ & + GRaWD (T=10) &\textbf{45.43} &\textbf{40.68} &\textbf{15.51} &\textbf{13.70}\\
% ~ & + GRaWD (T=10) &\textbf{45.55} &\textbf{40.7} &\textbf{15.5} &\textbf{13.7 }\\
\bottomrule
\end{tabular}}
\label{tb:ablation_zsl}

\vspace{3mm}
\vspace{0.8ex}
\caption{Zero-Shot Recognition from textual description on \textbf{CUB} and \textbf{NAB} datasets (Easy and Hard Splits) shwoing that adding GRaWD loss can improve the performance. \textit{tr} means the transductive setting.}
\label{tb:nab_cub}
\vspace{-1.0ex}
\resizebox{.57\textwidth}{!}{%
\begin{tabular}{@{}lccccccccc@{}}
\toprule
Metric & \multicolumn{4}{c}{Top-1 Accuracy (\%)} &  & \multicolumn{4}{c}{Seen-Unseen AUC (\%)} \\ \cmidrule(lr){2-5} \cmidrule(l){7-10} 
Dataset & \multicolumn{2}{c}{CUB} & \multicolumn{2}{c}{NAB} &  & \multicolumn{2}{c}{CUB} & \multicolumn{2}{c}{NAB} \\
Split-Mode & Easy & Hard & Easy & Hard &  & Easy & Hard & Easy & Hard \\ \midrule
% WAC-Linear \cite{elhoseiny2013write} & 27.0 & 5.0 & -- & -- &  & 23.9 & 4.9 & 23.5 & -- \\
% WAC-Kernel \cite{elhoseiny2016write} & 33.5 & 7.7 & 11.4 & 6.0 &  & 14.7 & 4.4 & 9.3 & 2.3 \\
% ESZSL \cite{romera2015embarrassingly} & 28.5 & 7.4 & 24.3 & 6.3 &  & 18.5 & 4.5 & 9.2 & 2.9 \\
ZSLNS \cite{Qiao2016} & 29.1 & 7.3 & 24.5 & 6.8 &  & 14.7 & 4.4 & 9.3 & 2.3 \\
SynC$_{fast}$ \cite{changpinyo2016synthesized} & 28.0 & 8.6 & 18.4 & 3.8 &  & 13.1 & 4.0 & 2.7 & 3.5 \\
% SynC$_{OVO}$ \cite{changpinyo2016synthesized} & 12.5 & 5.9 & -- & -- &  & 1.7 & 1.0 & 0.1 & --\\
ZSLPP \cite{Elhoseiny_2017_CVPR} &  37.2 & 9.7 &30.3 & 8.1 &  & 30.4 & 6.1 & 12.6 & 3.5 \\
FeatGen~\cite{xian2018feature} &  43.9 & 9.8 &36.2 & 8.7 &  & 34.1 & 7.4 & 21.3 & 5.6 \\ \midrule
LsrGAN (\textit{tr})~(Vyas et al. 2020) & 45.2 & 14.2 & 36.4 & 9.0 & & 39.5 & 12.1 & 23.2 & 6.4\\ 
 \quad +\textbf{GRaWD}& \textbf{45.6}\mathcolor{blue}{$^{+0.4}$} & 15.1\mathcolor{blue}{$^{+0.9}$} & 37.8\mathcolor{blue}{$^{+1.4}$} & 9.7\mathcolor{blue}{$^{+0.7}$} & & 39.9\mathcolor{blue}{$^{+0.4}$} & 13.3\mathcolor{blue}{$^{+1.2}$} & 24.5\mathcolor{blue}{$^{+1.3}$} & 6.7\mathcolor{blue}{$^{+0.3}$}\\ \midrule
%  \quad \quad +\textbf{SeGD} & 45.3\mathcolor{blue}{$^{+0.1}$} & 13.6\mathcolor{red}{$^{-0.6}$} & \textbf{38.2}\mathcolor{blue}{$^{+1.8}$} & 9.5\mathcolor{blue}{$^{+0.5}$} & & 39.6\mathcolor{blue}{$^{+0.1}$} & \textbf{12.8}\mathcolor{blue}{$^{+0.7}$} & \textbf{24.3}\mathcolor{blue}{$^{+1.1}$} & 6.4\mathcolor{blue}{$^{+0.0}$}\\ \midrule
GAZSL~\cite{Elhoseiny_2018_CVPR}& 43.7 & 10.3 & 35.6 & 8.6 &  & 35.4 & 8.7 & 20.4 & 5.8\\
\quad +{CIZSL}~\cite{elhoseiny2019creativity} &  44.6 & 14.4  & 36.6 \ & 9.3 &  & 39.2 & 11.9 & 24.5 & 6.4 \\ 
\quad + \textbf{GRaWD} &
 45.4\mathcolor{blue}{$^{+1.7}$}& \textbf{15.5} \mathcolor{blue}{$^{+5.2}$}  & \textbf{38.4} \mathcolor{blue}{$^{+2.8}$} & \textbf{10.1} \mathcolor{blue}{$^{+1.5}$} &  & \textbf{40.7} \mathcolor{blue}{$^{+5.3}$} & \textbf{13.7}\mathcolor{blue}{$^{+5.0}$} & \textbf{25.8}\mathcolor{blue}{$^{+5.4}$} & \textbf{7.4} \mathcolor{blue}{$^{+1.6}$} \\ 
%   \quad \quad +\textbf{SeGD} & \textbf{45.6}\mathcolor{blue}{$^{+1.9}$} & 14.9\mathcolor{blue}{$^{+4.6}$} & \textbf{39.7}\mathcolor{blue}{$^{+4.1}$} & \textbf{10.2}\mathcolor{blue}{$^{+1.6}$} & & \textbf{40.7}\mathcolor{blue}{$^{+5.3}$} & \textbf{13.7}\mathcolor{blue}{$^{+5.0}$} & \textbf{25.9}\mathcolor{blue}{$^{+5.5}$} & 7.2\mathcolor{blue}{$^{+1.4}$}\\ 
%  tf-VAEGAN & xx & xx & xx & xx &  & xx & xx & xx & xx\\
%  \quad + \textbf{GRaWD} & xx & xx & xx & xx & & xx & xx & xx & xx\\ \midrule
%  FeatGen & xx & xx & xx & xx &  & xx & xx & xx & xx\\
%   \quad + \textbf{GRaWD} & xx & xx & xx & xx & & xx & xx & xx & xx\\ \midrule
%  OOD & xx & xx & xx & xx & & xx & xx & xx & xx\\
%   \quad + \textbf{GRaWD} & xx & xx & xx & xx & & xx & xx & xx & xx\\ \midrule
%  Cycle-(U)WGAN & xx & xx & xx & xx & & xx & xx & xx & xx\\
%   \quad + \textbf{GRaWD} & xx & xx & xx & xx & & xx & xx & xx & xx\\ 
%  GAZSL~\cite{Elhoseiny_2018_CVPR}+{GRaWD + CRL only (Ours)} & 44.4 & 
%  LsrGAN (in) & -\\ 
%  \quad + {GRaWD (Ours)} & - \\ 
%  \quad + {GRaWD (Ours)}& - \\ \midrule
%  FA-SVAE~\cite{chen2020boundary} & -\\
% \quad +{GRaWD (Ours} & -\\
\bottomrule   
\vspace{-2mm}
\end{tabular}%
}
}  
\hfill
\parbox{.4\linewidth}{
% \centering
% \resizebox{.5\textwidth}{!}{
% \caption {Attribute based ZSL on AwA2, aPY and SUN. Compared with~\cite{8099557}.}
% \begin{tabular}{lccccccccc}
% \hline
%  & \multicolumn{3}{c}{AwA2} & \multicolumn{3}{c}{aPY} & \multicolumn{3}{c}{SUN} \\ \cmidrule(lr){2-4} \cmidrule(lr){5-7} \cmidrule(lr){8-10}
%  & H & S & U  & H & S & U  & H & S & U\\ \midrule
% GRaWT (T=0)$\sim [27]$               & 32.3 & 80.5 & 20.2 & 23.0 & 78.9 & 13.4 & 26.0 & 31.6 & 22.2\\
% GRaWT (T=3) & 31.6 & 80.7 & 19.7 & 22.4 & 75.8 & 13.1 & 25.8 & 31.1 & 22.1\\ \hline
% \textbf{GRaWD}        & \textbf{39.0} & \textbf{88.3} & \textbf{25.0} & \textbf{27.2} & \textbf{83.2} & \textbf{16.3} & \textbf{27.9} & \textbf{37.3} & \textbf{22.3}\\
% \hline
% \end{tabular}
% }.       
\vspace{-4.5mm}
\caption{Attribute based ZSL on AwA2, aPY and SUN. Compared with Haeusser et al. (2017).} 
\vspace{-1.5ex}
\centering
\resizebox{.40\textwidth}{!}{
\begin{tabular}{lccccccccc}
\hline
 & \multicolumn{3}{c}{AwA2} & \multicolumn{3}{c}{aPY} & \multicolumn{3}{c}{SUN} \\ \cmidrule(lr){2-4} \cmidrule(lr){5-7} \cmidrule(lr){8-10}
 & H & S & U  & H & S & U  & H & S & U\\ \midrule
GRaWT (T=0)               & 32.3 & 80.5 & 20.2 & 23.0 & 78.9 & 13.4 & 26.0 & 31.6 & 22.2\\
GRaWT (T=3) & 31.6 & 80.7 & 19.7 & 22.4 & 75.8 & 13.1 & 25.8 & 31.1 & 22.1\\ \hline
\textbf{GRaWD}        & \textbf{39.0} & \textbf{88.3} & \textbf{25.0} & \textbf{27.2} & \textbf{83.2} & \textbf{16.3} & \textbf{27.9} & \textbf{37.3} & \textbf{22.3}\\
\hline
\end{tabular}}
\label{tb:ablation_zsl_attributes}

\vspace{3mm}
\caption{Zero-Shot Recognition on class-level attributes of \textbf{AwA2}, \textbf{aPY} and \textbf{SUN} datasets, showing that GRaWD loss can improve the performance on attribute-based datasets. }
\vspace{-2.0mm}
\resizebox{.45\textwidth}{!}{%
\begin{tabular}{@{}lccccccc@{}}
\toprule
 & \multicolumn{3}{c}{Top-1 Accuracy(\%)} &  & \multicolumn{3}{c}{Seen-Unseen H} \\ \cmidrule(lr){2-4} \cmidrule(l){6-8} 
 & AwA2 & aPY & SUN &  & AwA2 & aPY & SUN \\ \midrule
% DAP~\cite{Lampert2014} & 46.1 & 33.8 & 39.9 &  & -- & 9.0 & 7.2 \\
% SSE~\cite{zhang2015zero} & 61.0 & 34.0 & 51.5 &  & 14.8 & 0.4 & 4.0 \\
SJE~\cite{akata2015evaluation} & 61.9 & 35.2 & 53.7 &  & 14.4 & 6.9 & 19.8 \\
% ESZSL~\cite{romera2015embarrassingly} & 58.6 & 38.3 & 54.5 &  & 11.0 & 4.6 & 15.8 \\
LATEM~\cite{xian2016latent} & 55.8 & 35.2 & 55.3 &  & 20.0 & 0.2 & 19.5 \\
ALE~\cite{akata2016label} & 62.5 & 39.7 & 58.1 &  & 23.9 & 8.7 & 26.3 \\
% CONSE~\cite{norouzi2013zero} & 44.5 & 26.9 & 38.8 &  & 1.0 & -- & 11.6 \\
SYNC~\cite{changpinyo2016synthesized} & 46.6 & 23.9 & 56.3 &  & 18.0 & 13.3 & 13.4 \\
SAE~\cite{kodirov2017semantic} & 54.1 & 8.3 & 40.3 &  & 2.2 & 0.9 & 11.8 \\
DEM~\cite{zhang2016learning} & 67.1 & 35.0 & 61.9 &  & 25.1 & 19.4 & 25.6 \\
% DEVISE~\cite{frome2013devise} & 59.7 & 39.8 & 56.5 &  & {27.8} & 9.2 & 20.9 \\  
FeatGen~\cite{xian2018feature} & 54.3 & 42.6 & 60.8 &  & 17.6 & 21.4 & 24.9 \\
% \quad + {CIZSL}~\cite{elhoseiny2019creativity} & 60.1  & 43.8 & 59.4 &  & 19.1 & 24.0  & 26.5  \\
% \midrule
cycle-(U)WGAN~\cite{felix2018multi} & 56.2 & 44.6 & 60.3 &  & 19.2 & 23.6 & 24.4 \\ \midrule
% \quad + {CIZSL}~\cite{elhoseiny2019creativity} & 63.6 & {45.1}   & {64.2}   &  & 23.9   & {26.2}   & 27.6 \\ \midrule
LsrGAN (\textit{tr})~(Vyas et al. 2020) & $60.1^*$ & $34.6^*$ & 62.5 & & $48.7^*$ & $31.5^*$ & 44.8\\ 
% %  \color{blue}LsrGAN (Transductive, Rep) & - & - & 62.01 & - & - & 44.6\\ 
 \quad + {\textbf{GRaWD}}& 63.7\mathcolor{blue}{$^{+3.6}$} & 35.5\mathcolor{blue}{$^{+0.9}$} & \textbf{64.2}\mathcolor{blue}{$^{+1.7}$} & & \textbf{49.2}\mathcolor{blue}{$^{+0.5}$} & \textbf{32.7}\mathcolor{blue}{$^{+1.2}$} & \textbf{46.1}\mathcolor{blue}{$^{+1.3}$} \\ \midrule
%  \quad \quad + \textbf{SeGD}& 61.5\mathcolor{blue}{$^{+1.4}$} & 35.2\mathcolor{blue}{$^{+0.6}$} & 63.1\mathcolor{blue}{$^{+0.6}$} & & 46.3\mathcolor{red}{$^{-2.4}$} & 31.0\mathcolor{red}{$^{-0.5}$} & 45.6\mathcolor{blue}{$^{+0.8}$} \\ \midrule
%  LsrGAN (in) & -\\   
% \quad + {GRaWD (Ours)}& - \\ \midrule
% FA-SVAE~\cite{chen2020boundary}& -\\
% \quad +{GRaWD (Ours} & -\\
%  tf-VAEGAN~\cite{narayan2020latent} & 72.2 & 38.4* & 66.0 & & 66.6 & 32.1* & 43.0\\
%  \quad + \textbf{GRaWD} & 71.8\mathcolor{red}{$^{-0.4}$} & 39.6\mathcolor{blue}{$^{+1.2}$} & 66.5\mathcolor{blue}{$^{+0.5}$} & & 67.4 \mathcolor{blue}{$^{+0.8}$}  & 32.9 \mathcolor{blue}{$^{+0.8}$}  & 43.0 \mathcolor{blue}{$^{+0.0}$} \\ \midrule
 
GAZSL~\cite{Elhoseiny_2018_CVPR} & 58.9 & 41.1 & 61.3 &  & 15.4 & 24.0 & 26.7 \\ 
\quad + {CIZSL}~\cite{elhoseiny2019creativity} & {67.8} & {42.1}& 63.7  &  & 24.6 & {25.7} & {27.8}  \\
\quad + {\textbf{GRaWD}} & \textbf{68.4}\mathcolor{blue}{$^{+9.5}$} & \textbf{43.3}\mathcolor{blue}{$^{+2.2}$} & 62.1\mathcolor{blue}{$^{+0.8}$} &  & 39.0\mathcolor{blue}{$^{+23.6}$} & 27.2\mathcolor{blue}{$^{+3.2}$} & 27.9\mathcolor{blue}{$^{+1.2}$} \\ 
%  \quad \quad + \textbf{SeGD}& \textbf{68.7}\mathcolor{blue}{$^{+9.8}$} & 42.8\mathcolor{blue}{$^{+1.7}$} & 62.9\mathcolor{blue}{$^{+1.6}$} & & \textbf{40.1}\mathcolor{blue}{$^{+24.7}$} & \textbf{27.7}\mathcolor{blue}{$^{+3.7}$} & \textbf{28.9}\mathcolor{blue}{$^{+2.2}$}\\ 
% \quad \quad +{SGC (Ours)} & {68.7}\mathcolor{blue}{$^{+9.8}$} & 42.8 & 62.9 & & \textbf{40.13} & \textbf{27.66} & \textbf{28.92}\\ \midrule
%  FeatGen & xx & xx & xx & & xx & xx & xx\\
%   \quad + \textbf{GRaWD} & xx & xx & xx & & xx & xx & xx\\ 
%  OOD (reported)& - & - & - & & 70.3 & - & 71.9\\
%  OOD (reproduced)& - & - & - & & - & - & -\\
%   \quad + \textbf{GRaWD} & xx & xx & xx & & xx & xx & xx\\ \midrule
%  cycle-(U)WGAN~\cite{felix2018multi} & xx & xx & xx & & xx & xx & xx\\
%   \quad + \textbf{GRaWD} & xx & xx & xx & & xx & xx & xx\\
\bottomrule
\end{tabular}       
}
\label{tb:awa2apysun}
\vspace{-3.5mm}
}
\vspace{-2.0ex}
\end{table*}

\section{Experiments}

\subsection{Purely Inductive  Generative ZSL Experiments}
% \vspace{-1mm}
% We performed our experiments on existing zero-shot learning benchmarks with both text descriptions and attributes as semantic class descriptions. Text-description setting is more challenging  since comes at the class level and are extracted Wikipedia which are noisier.   
% We found random walk steps $T$  easy to tune based on validation set and use discounting factor $\gamma=0.7$ in all our experiments. More implementation details can be found in the supplementary.  %For our experiments, we perform cross-validation to obtain the value of $\lambda$ that varies by setting.
      
\noindent \emph{Purely Inductive Evaluation in ZSL:} 
Our focus in this paper is to learn a good representation of unseen visual spaces without accessing any unseen class information during training. However, most recent papers jointly train an extra classifier (e.g., an MLP) with their proposed generative model~\cite{narayan2020latent, han2021contrastive}. More concretely,  this classifier is trained with generated $\tilde{X}_u = G(u_k, z)$ from unseen semantic description $u_k$. While in GZSL, training seen images $X_s$ along with $\tilde{X}_u$ is introduced as input of the extra classifier.  We refer to methods that assume access to unseen class descriptions during training as semantic transductive ZSL (even if not using unlabeled images of unseen classes). Accessing unseen information before evaluation is not in line with our focus on learning generative unseen learning representation. This is also less realistic if we aim at purely inductive zero-shot learning. Following purely inductive ZSL settings  (e.g., ~\cite{Elhoseiny_2018_CVPR, elhoseiny2019creativity}), we use NN-classification on the generated features for evaluation, which avoids accessing any unseen semantic information before testing.

% We performed our experiments on existing zero-shot learning benchmarks with text descriptions and attributes as semantic class descriptions. The text-description setting is more challenging since it comes at the class level and is extracted from Wikipedia, which is noisier. 
% We found random walk steps $T$  easy to tune based on the validation set and used discounting factor $\gamma=0.7$ in all our experiments. We showed that GRaWD could improve on multiple methods for most text-based and attribute-based datasets.

We performed experiments on existing ZSL benchmarks with text descriptions and attributes as semantic class descriptions. Note that the text description setting is more challenging because it is at the class-level and is extracted from Wikipedia, which is noisier.   
We found that random walk steps $T$  easy to tune using  the validation set. % and we used discounting factor $\gamma=0.7$ in all experiments. %We demonstrate that the proposed GRaWD loss outperforms multiple methods on most text-based and attribute-based datasets. 

\noindent \textbf{Text-Based ZSL.}
We performed our text-based ZSL  experiments on Caltech  UCSD Birds-2011 (CUB)~\cite{WahCUB_200_2011} containing 200 classes with 11, 788 images and North America Birds (NAB)~\cite{Horn2015} which has 1011 classes with 48, 562 images.
We use two metrics widely used in evaluating ZSL recognition performance: standard zero-shot recognition with the Top-1 unseen class accuracy and Seen-Unseen Generalized Zero-shot performance with Area under Seen-Unseen curve~\cite{chao2016empirical}. 
We follow ~\cite{chao2016empirical, Elhoseiny_2018_CVPR,elhoseiny2019creativity} in using the Area Under SUC to evaluate the generalization capability of class-level text zero-shot recognition on four splits (CUB Easy, CUB Hard, NAB Easy, and NAB Hard). The hard splits are constructed such that unseen bird classes from super-categories do not overlap with seen classes. Our proposed loss function improves over older methods on all datasets on both Easy and SCE(hard) splits, as shown in Table~\ref{tb:nab_cub}. We show improvements in the range of 0.8-1.8\%  Top-1 accuracy. We also show improvements in AUC, ranging from 1-1.8\%. From Table~\ref{tb:nab_cub}, we show that GAZSL~\cite{Elhoseiny_2018_CVPR}+GRaWD has an average relative Seen-Uneen AUC improvement over GAZSL~\cite{Elhoseiny_2018_CVPR}+CIZSL~\cite{elhoseiny2019creativity} and GAZSL~\cite{Elhoseiny_2018_CVPR} only of 9.29\% and 30.89\%. We achieved SOTA results for text datasets. In Table~\ref{tb:ablation_zsl}, we performed an ablation study where we show that longer random walks performed better hence giving higher accuracies and AUC scores for both easy and hard split for CUB Dataset. With longer walks, the model could have a more holistic view of the generated visual representation in a way that enables better deviation of unseen classes from unseen classes. Therefore, we used T=10 for our experiments.

% \kai{\emph{Inductive ZSL:} we showed that GAZSL[76]+ GRaWD  has an average relative improvement over GAZSL[76]+CIZSL[14]  and GAZSL[76] of  24.92\% and  61.35\% (H)  on attribute datasets and   9.29\% and  30.89\% (AUC)  on text-based datasets, respectively; computed from Table 2 \& 3 in the paper. We achieved SOTA results for text datasets.}

% \kai{\emph{Transductive ZSL:} Despite that  our loss  does not use unseen class descriptors, it can still improve on average on [64 from ECCV2020] (transductive) by 1.96\%  on attribute datasets and 2.91\% on text-based datasets.}

% \kai{\textbf{SeGD marginal improvement.}  SeGD is not our main contribution and is used on top of  GRaWD to improve the performance  (e.g., from 39.0\% to 40.1\% on AwA2 and 27.9\% to 28.9\% on SUN).  In transductive setting, SeGD is not as helpful as in inductive setting (more realistic).}

\noindent \textbf{Attribute-Based ZSL.} We performed experiments on the widely used GBU~\cite{gbu} setup, where we use class attributes as semantic descriptors. We performed these experiments on the AwA2~\cite{lampert2009learning}, aPY~\cite{farhadi2009describing}, and SUN~\cite{patterson2012sun} datasets. 
In Table~\ref{tb:awa2apysun}, we see that GRaWD outperforms all of the existing methods on seen-unseen harmonic mean for AwA2, aPY, and SUN datasets. In the case of the AwA2 dataset, it outperformed the compared method by a significant margin, i.e., 15.1\%. It is also competent with  existing methods in Top-1 accuracy while improving on AwA2 4.8\%. From Table~\ref{tb:awa2apysun},  GAZSL~\cite{Elhoseiny_2018_CVPR}+GRaWD has an average relative improvement  over GAZSL~\cite{Elhoseiny_2018_CVPR}+CIZSL~\cite{elhoseiny2019creativity} and GAZSL~\cite{Elhoseiny_2018_CVPR} of 24.92\% and 61.35\% in harmonic mean.

Table \ref{tb:ablation_zsl} and \ref{tb:ablation_zsl_attributes} show that deviation signal in GRaWD is critical to achieve better performance since the calculated metrics are much better for GRaWD compared to GRaWT for both text-based and attribute based ZSL. The performance can severely degrade without the deviation signal. % Bottom section of Tab.~\ref{tb:ablation_zsl} shows that longer walk lengths benefits the training as model is able to explore larger section of unseen representations. 
%\kai{  
Tab.~\ref{tb:ablation_zsl} (bottom section) shows that longer walk lengths benefits the training as model is encouraged to globally explore larger segments of unseen representations' manifold.%}

\noindent \textbf{GRaWD Loss for Transductive ZSL.} 
We also apply our GRaWD loss to transductive ZSL setting and choose choose LsrGAN~\cite{vyas2020leveraging} as the baseline model. Our loss can also improve LsrGAN on both text-based datasets and attribute-based datasets on most metrics ranging from 0.3\%-3.6\%.  Despite that  our loss  does not use unseen class descriptors, it can still improve on average on LsrGAN (transductive) by 1.96\%  on attribute datasets and 2.91\% on text-based datasets. However, in line with our expectations, the former improvement in the purely inductive setting is more significant.

\subsection{Novel Art Generation Experiments.}
\label{sec:art_Experiments}

 We performed our Art experiments on the WikiArt dataset, containing 81k images of 27 different  styles~\cite{wikiart20}.%,can_2017}. 
% We used  WikiArt dataset which contains 81k images of 27 different art styles~\cite{wikiart20,can_2017}. We trained our models on two architectures namely StyleGAN~\cite{Karras_2019} and DCGAN~\cite{radford2015} to generate art from random noise unconditioned on nothing. We use DCGAN~\cite{radford2015}, StyleGAN~\cite{Karras_2019} and {StyleGAN2}~\cite{karras2020analyzing}. The training details for our method can be found in the supplementary section. 

\begin{table}[!b]
\vspace{-3.0mm}
\centering
\caption{Human experiments on generated art from vanilla GAN, GRaWD, and CAN losses. Models trained on GRaWD obtained the highest mean likeability in all the groups. More people believed the generated art to be real for the artwork generated from model trained using GRaWD.}
\label{tb:likeability}
\vspace{-2mm}
\begin{adjustbox}{width=8.3cm}
\begin{tabular}{@{}p{2.5cm}cccccccccc@{}}
\toprule
% \multicolumn{8}{c}{Generated Image Likeability} \\ \midrule
& & \multicolumn{5}{c}{Likeability Mean}  & Turing Test \\ \cmidrule(lr){3-7} \cmidrule(lr){8-8}
Loss & Architecture  & Q1-mean(std)    &  NN $\uparrow$ & NN $\downarrow$ & Entropy $\uparrow$ & Random   & Q2(\% Artist) \\ \midrule
CAN~\cite{elgammal2017can} & DCGAN  & 3.20(1.50) & - & - & - & - & 53 \\ \midrule
GAN (Vanilla)   & StyleGAN & 3.12(0.58)    &  3.07        & 3.36          & 3.00             & 3.06      & 55.33  \\
CAN & StyleGAN     & 3.20(0.62)      &  3.01         & 3.61          & 3.05             & 3.11  & 56.55   \\
\textbf{RW-T3 (Ours)} & StyleGAN & \textbf{3.29(0.59)}     & 3.05         & 3.58          & 3.13             & \textbf{3.38}   & 54.08   \\ 
\textbf{RW-T10 (Ours)} & StyleGAN  & \textbf{3.29(0.63)}      & \textbf{3.15}         & \textbf{3.67}          & \textbf{3.15}             & 3.17  & \textbf{58.63}   \\ \midrule

GAN (Vanilla) & StyleGAN2  & 3.02(1.15)               & 2.89 & 3.30 & 2.79 & 3.09 & 54.01 \\ 
CAN  & StyleGAN2  & 3.23(1.16)                        & 3.27 & 3.34 & 3.11 & 3.21 & 57.9 \\ 
\textbf{RW-T3 (Ours)} & StyleGAN2 & \textbf{3.40(1.1)} & \textbf{3.30} & \textbf{3.61} & \textbf{3.33} & \textbf{3.35} & \textbf{64.0} \\ 

% \midrule
% Abstract art~\cite{elgammal2017can} & Human  & 3.3(0.43) & N/A & N/A & N/A & N/A & 85 \\ 
% Art Basel ~\cite{elgammal2017can} & Human  & 2.8(0.6)  & N/A & N/A & N/A & N/A & 41 \\ 
% Artist sets ~\cite{elgammal2017can} & Human & 3.1(0.63) & N/A & N/A & N/A & N/A & 62 \\ 
\midrule
\end{tabular}
\end{adjustbox}
\vspace{-4.0mm}
\end{table}

\begin{table}[!b]
\vspace{-3.0mm}
\centering
\caption{
Normalized mean ranking (lower the better) calculated from the likeability experiment. We take the mean rating of each artwork on both CAN and GRaWD losses. We then stack, sort, normalize them to compute the normalized mean rank. The numbers are corresponding normalized ranks from the models in the row above them.}
\label{tb:ranking_st2_st1}
\vspace{-2mm}
\begin{adjustbox}{width=8cm}
\begin{tabular}{lrrrr}\toprule
&\multicolumn{3}{c}{Normalized Mean Ranks} \\\midrule
 & CAN/\textbf{RW-T10} & CAN/\textbf{RW-T3} & CAN/\textbf{RW-T10}/\textbf{RW-T3}\\
\cmidrule{2-4}
StyleGAN1 & 0.53/\textbf{0.47} & 0.53/\textbf{0.47} & 0.52/\textbf{0.48}/\textbf{0.50} \\
\midrule
&CAN/\textbf{RW-T3} &GAN/\textbf{RW-T3}  & CAN/GAN/\textbf{RW-T3}  \\
\cmidrule{2-4}
StyleGAN2 & 0.54/\textbf{0.46} & 0.59/\textbf{0.41} & 0.49/0.59/\textbf{0.42} \\

% & \textbf{RW-T3} & \textbf{RW-T3} & vs \textbf{RW-T3} \\
\bottomrule
\end{tabular}
\end{adjustbox}
\end{table}

\noindent \textbf{Baselines.}
We performed comparisons with two baselines, i.e., (1) the vanilla  GAN for the chosen architecture, and (2) adding Holistic-CAN loss~\cite{sbai2018design} (an improved version of CAN~\cite{elgammal2017can}). For simplicity, we refer the Holistic-CAN as CAN. 
   
\begin{figure}[t!]
  \vspace{-2mm}
  \centering
  %\begin{subfigure}[b]{0.4\linewidth}
%   \includegraphics[width=9cm]{figures/most_liked.png}
    \includegraphics[width=8.0cm]{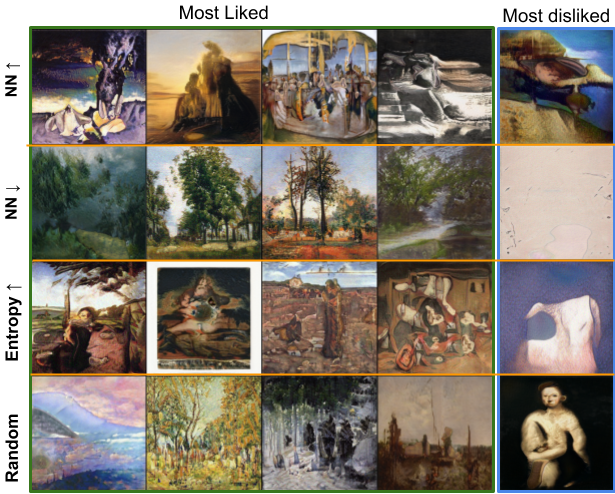}
\vspace{-3mm}
  %\end{subfigure}
  \caption{Most liked and disliked art generated using StyleGAN trained on GRaWD for the different groups.%  model trained on GRaWD loss for the different groups.
  }
  \label{fig:most_least_liked}
  \vspace{-3mm}
\end{figure}

\noindent \textbf{Nomenclature.}
 Here, the models are referred as RW-T(value), where RW means GRaWD loss, and T is the number of steps. We name our models according to this convention throughout. We perform human subject experiments to evaluate generated art. We set value of the loss coefficient $\lambda$ as 10. We divide the generations from these models into four groups, each containing 100 images; see examples in Fig.~\ref{fig:most_least_liked}.
  \begin{itemize}
  \item{
\noindent \textbf{NN$\uparrow$.}  Images with high nearest neighbor (NN)  distance from the training dataset.} 
% We performed comparisons with  two baselines (1) the standard  GAN for the chosen architecture. (2) adding Holistic-CAN loss~\cite{sbai2018design} (an improved version of CAN~\cite{elgammal2017can}). For simplicity, we refer to it as CAN. The value of the loss coefficient, $\lambda$ is 10 unless specified. The models are named as RW-T(value), where RW means GRaWD loss, and T is the number of steps. We name our models according to this specific method throughout this section.\\
% We also perform human subject experiments to evaluate generated art. We divide the generations from these models into four groups, each containing 100 images.\\
% \noindent \textbf{NN$\uparrow$.}  Images with high nearest neighbour (NN)  distance from the training dataset. 
\item{\noindent \textbf{NN$\downarrow$.} Images with low NN distance from the training  dataset. }
\item{\noindent \textbf{Entropy $\uparrow$.} Images with high entropy of the probabilities from a style classifier. }
\item{\noindent\textbf{Random (R).} A set of random images.}
\end{itemize}
%\noindent We use this nomenclature to refer to these groups. 
\noindent\label{par:nomenclature}For example, we denote generations using our GRaWD loss with $T$=10, and NN$\uparrow$ group as RW-T10\_NN$\uparrow$. Fig.~\ref{fig:most_least_liked} shows top liked/disliked paintings according to  human evaluation.

\begin{figure}[b!]
    \centering
    \includegraphics[width=8cm]{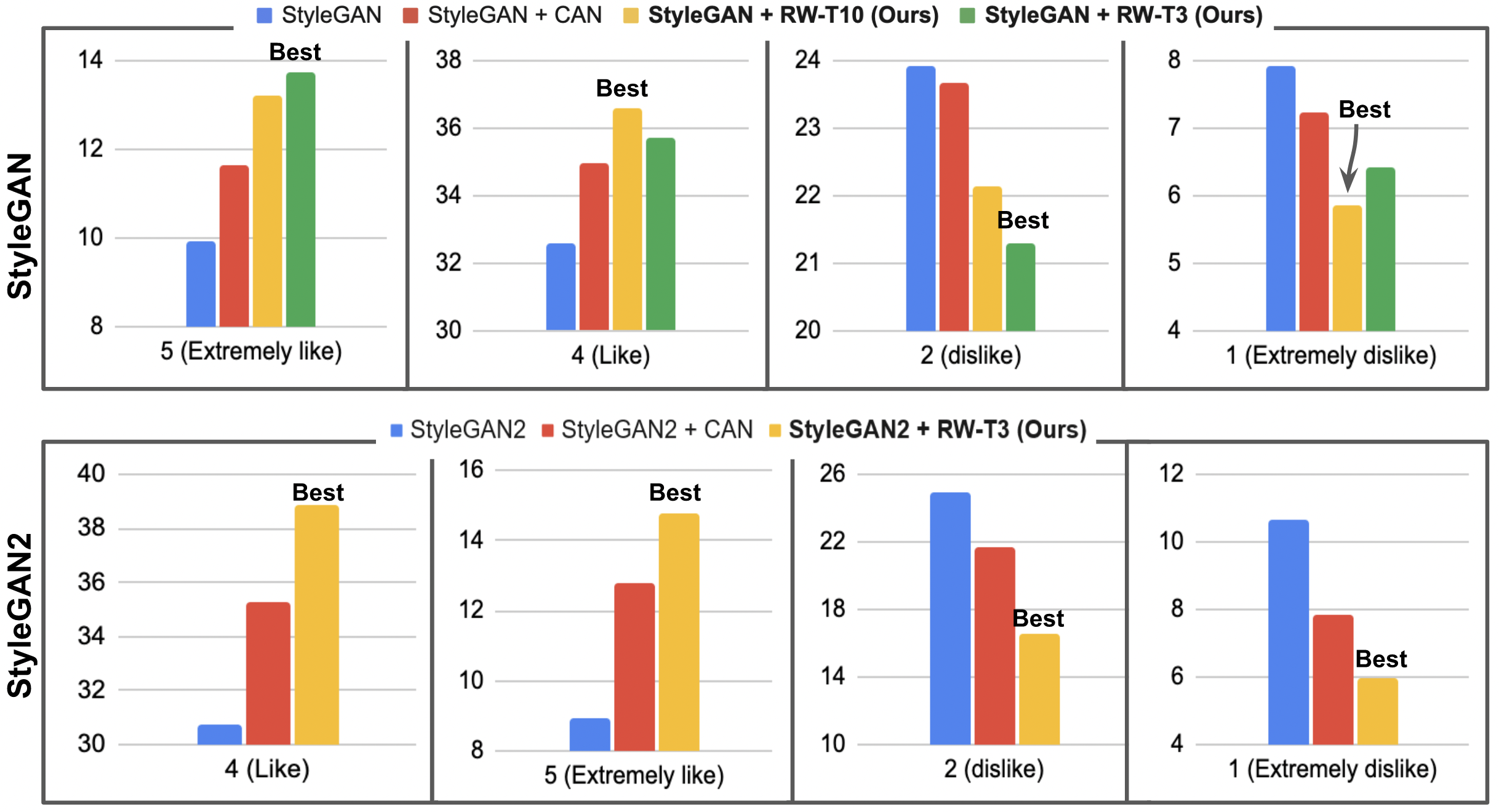}
    \vspace{-2mm}
    \caption{Percentage of each rating from human subject experiments on generated images.  Compared to CAN, images from our loss are rated (5,4) by a significantly larger share of people, and are rated (1,2) by fewer people.}
    \label{fig:human-subject-percentage}
    \vspace{-4mm}
\end{figure}

\noindent \textbf{Human Evaluation}.% Experiments.} 
 We performed our human subject MTurk experiments based on StyleGAN1\cite{Karras_2019} \& StyleGAN2 \cite{karras2020analyzing} architecture's vanilla, CAN, and GRaWD variants.
We divide the generations into four groups described above.  
We collect five responses for each art piece (400 images), totaling 2000 responses per model by 341 unique workers. We asked people to rate generations from 1 (extremely dislike) to 5 (extremely like), which was the first question (Q1). In Q2, we asked if a computer or an artist generates the images (Turing Test). More setup details are in the supplementary.
 We found that art from the trained StyleGAN1 and StyleGAN2 on our loss were more likeable and more people believed them to be real art, as shown in Table~\ref{tb:likeability}. 
 {For StyleGAN1, adding GRaWD loss resulted in  38\% and 18\% more people giving a full rating of 5 over vanilla StyleGAN1 and StyleGAN1 + CAN loss, respectively, see Fig.~\ref{fig:human-subject-percentage}. For StyleGAN2, these improvements were 65\% and 15\%}.
 Table~\ref{tb:ranking_st2_st1} shows that images from the StyleGAN model on our loss have a better rank when combined with other sets of the table. 
% We compared generated art of the trained StyleGAN 1 and 2 model on GRaWD and CAN losses. For a pair of images, people submitted their preference. We collected 5 responses each for 600 pairs of art by 300 unique workers. Table~\ref{tb:preference} shows that our loss is preferred more.

\begin{figure}[t!]
    \centering
    \includegraphics[width=8.0cm]{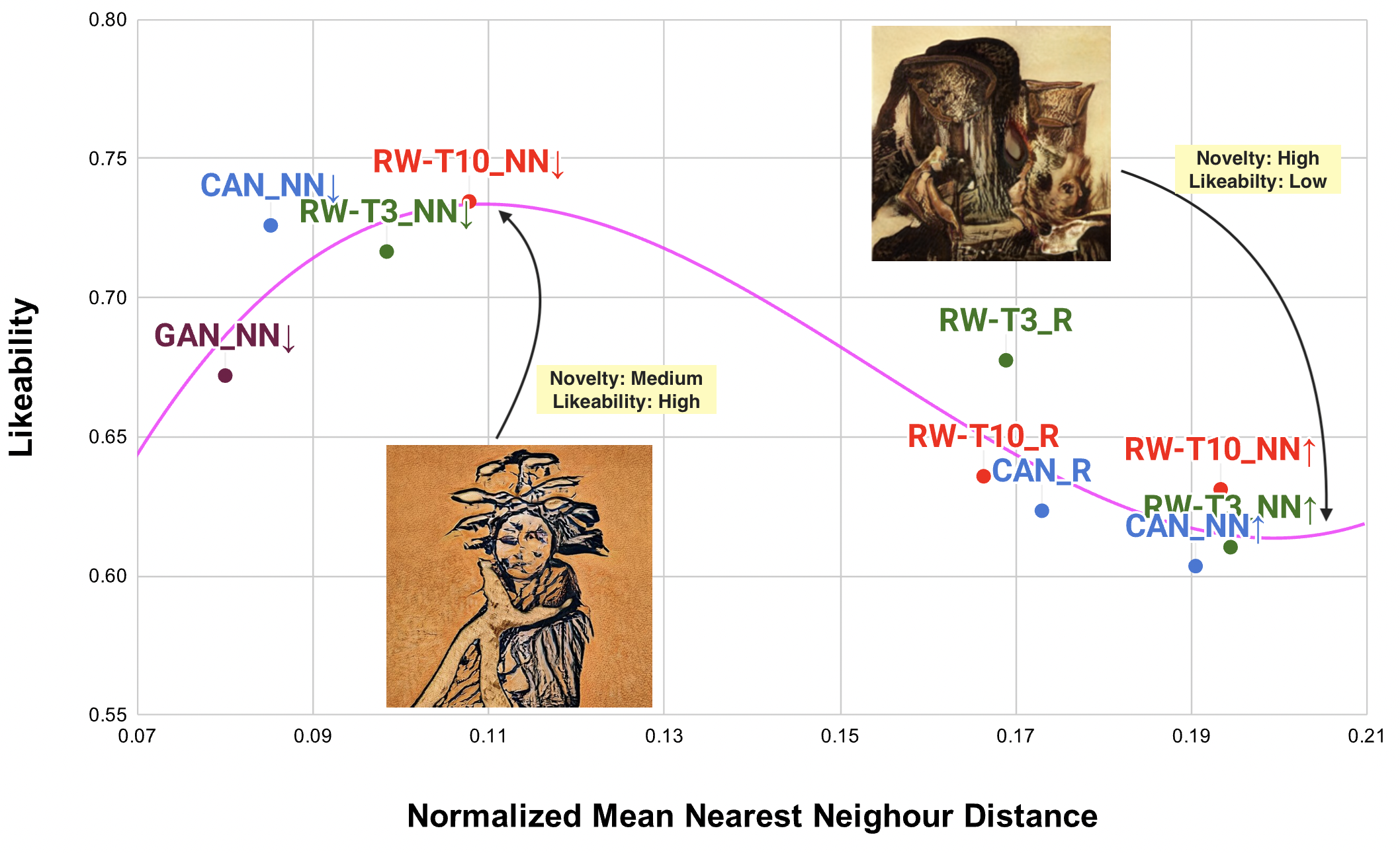}
    		\vspace{-2mm}
    \captionof{figure}{Empirical approximation of Wundt Curve ~\cite{packard1975aesthetics, wundt1874grundzuge}.
 The color of the data point represents a specific model and its label specifies the group named according to nomenclature. Art from the NN $\uparrow$  group has lower likeability than the NN $\downarrow$ group. Examples of a high and low likeability artwork and its novelty are shown.} 
    \label{fig:wundt}   
    \vspace{-4mm}
\end{figure}

% \begin{table}[b!]
%     \centering
%     % \vspace{-3mm}
%     \caption{Evaluator preference percentage for generated images for both GRaWD and CAN loss on the StyleGAN architecture. We split the preferred images into two groups based on their NN distance, and then the preference percentage is calculated for these groups.}
%     % \vspace{-2mm}
%     \label{tb:preference}      
%     \begin{adjustbox}{width=8cm}
%     \begin{tabular}{lrrrr}\toprule
%     % &\multicolumn{3}{c}{Normalized Mean Ranks} \\
%     % \cmidrule{2-4}
%     & Architecture &Low NN distance split  &High NN distance split  \\
%     \midrule
%     CAN & StyleGAN1 & 0.46 &0.48 \\
%     \textbf{RW-T10} & StyleGAN1 & \textbf{0.54} &\textbf{0.52} \\
%     \midrule
%     CAN & StyleGAN2 &0.46 &0.43 \\
%     \textbf{RW-T10} & StyleGAN2 &\textbf{0.54} &\textbf{0.56}  \\   
%     \bottomrule
%     \end{tabular}
%     \end{adjustbox}
%     % \vspace{-0.5ex}
  
%     % \includegraphics[width=6cm]{figures/preference.png}
%     % \caption{Evaluator preference percentage for generated images for both GRaWD and CAN loss on StyleGAN architecture. The preferred images were split into two groups based on their NN distance and then the preference percentage is calculated on these groups.}
%     % \label{fig:preference}
%     % \vspace{-3mm}
% \end{table}

\begin{figure}[b!]
\centering
% \vspace{-4.5mm}
\includegraphics[width=8.0cm]{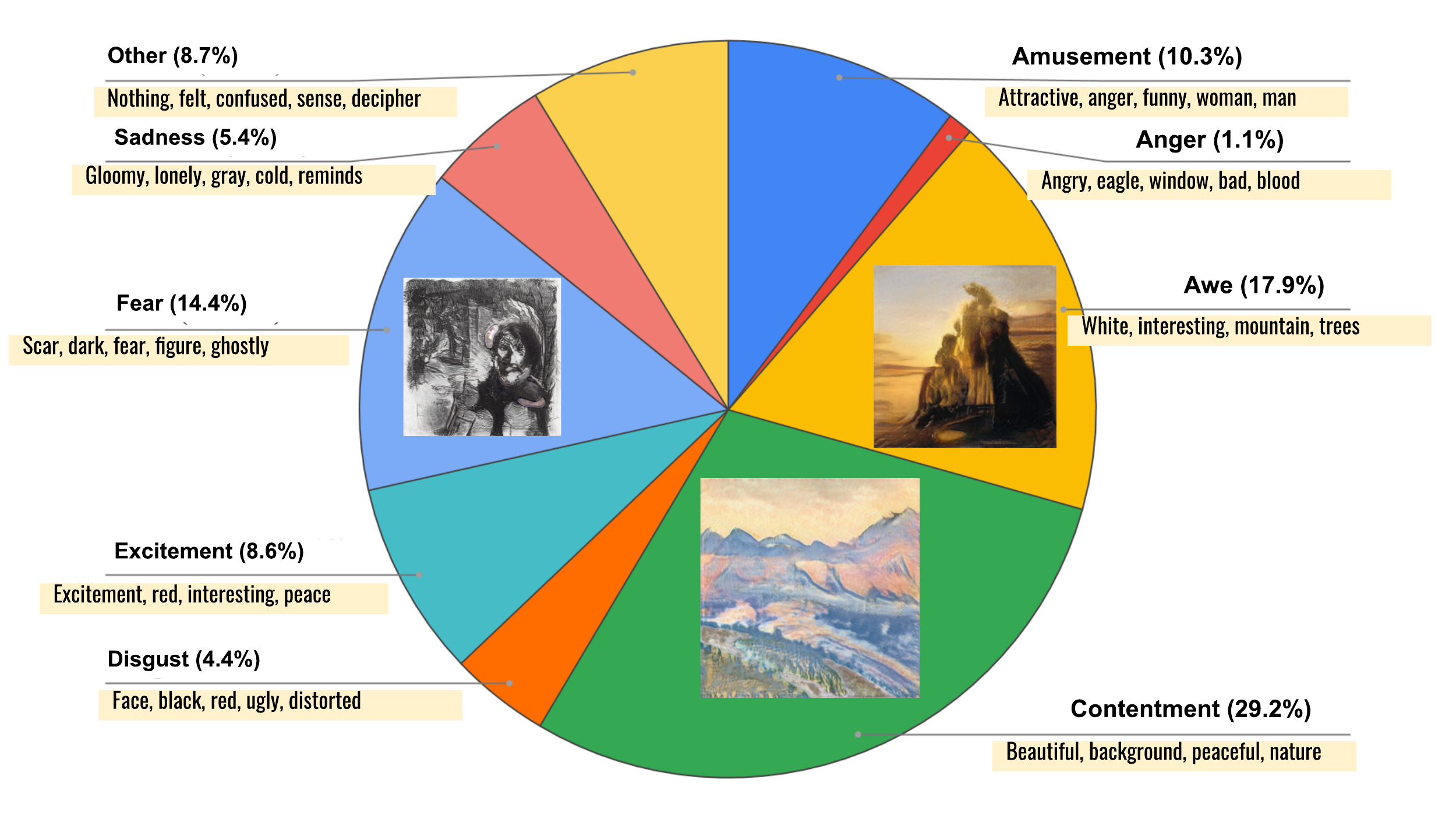}
\vspace{-2mm}
\captionof{figure}{Distribution of emotional responses for StyleGAN generated art trained on GRAWD loss. An example image for fear, awe, and contentment is shown. The box beneath for each emotion shows the most frequent words used by evaluators to describe their feeling. These responses were collected from a survey on Mechanical Turk. }%\kai{According to the rebuttal: ``Is Fig7 correct? Yes, it is. We generated only pie chart first and later edited the image to fill the missing parts (e.g, words). We will improve it in the final version.''}.} %gathered by a survey to collect emotional responses and describe in text. }
\label{fig:emotions}
\vspace{-3mm}
\end{figure}

% \begin{figure}[hb!]
% \vspace{-2.5mm}
% \includegraphics[width=8cm]{figures/mother_nature.png}
% \vspace{-2mm}
% \caption{``Mother Nature Needs Help'', our model generated painting on the far left. The painting is followed by its Nearest neighbors  from Wikiart dataset based on ResNet18 features~\cite{he2016resnet}.  The NN images are quite different in content indicates that  ``Mother Nature Needs Help'' is novel.}
% \label{fig_mother_nature}
% \vspace{-1.5mm}
% \end{figure}

\noindent \textbf{Wundt Curve Analysis.} We approximate the Wundt curve~\cite{packard1975aesthetics, wundt1874grundzuge} by fitting a degree 3 polynomial on  a scatter plot of  normalized likeability vs. mean NN distance ( novelty measure).  Generations are more likable if the deviation from existing art is moderate but not too much; see Fig.~\ref{fig:wundt}. Compared to CAN and GAN, our loss achieves on balance novel images that are more preferred.% if this deviation is large, people tend to dislike. 

%  It empirically shows that people like if models generate images that are similar to the existing art. If something too novel (high NN distance) is generated, people dislike them.\\
% Figure \ref{fig:most_least_liked} shows views of nature, mostly trees. While the high nearest neighbour (left, row 1) has some vivid colors but people can develop their own interpretation of the art.
\noindent \textbf{Emotion Experiments.} Evaluators selected the emotion they felt after looking at each art and justified their chosen emotion in text. We collected 5 responses each for a set of 600 generated artworks from 260 unique workers. Fig.~\ref{fig:emotions} shows the distribution over the opted emotions, which are diverse but mostly positive. However, some generations construct negative emotions like fear. Fig.~\ref{fig:emotions} also shows the most frequent words for each emotion after removing stop words. Notable positive words include ``funny", ``beautiful", ``attractive",  and negative words include ``dark", ``ghostly" which are associated with feelings like fear and disgust.

\section{Conclusion}
\vspace{-0.5mm}
%We propose Generative Random Walk Deviation (GRaWD) loss and showed in our experiments that it improves the capability of generative models to learn better representation for understanding unseen classes on several zero-shot learning benchmarks and to generate novel visual content trained on WikiArt dataset. We think the improvement is due to the more global nature of our learning mechanism which operates at the  minibatch level producing generations that are message-passing to each other to facilitate better deviation of unseen classes/styles from seen ones

We propose Generative Random Walk Deviation (GRaWD) loss and showed that it improves generative models' capability to better understand unseen classes on several zero-shot learning benchmarks and generate novel visual content trained on WikiArt dataset. We think the improvement is due to our learning mechanism's global nature, which operates at the minibatch level producing generations that are message-passing to each other to facilitate better deviation of unseen classes/styles from seen ones.

% {\small
% % \bibliographystyle{aaai22}
% \bibliography{aaai22}
% }

\nobibliography*

% Use \bibliography{yourbibfile} instead or the References section will not appear in your paper
  
% \newpage
{
\small
\bibliography{aaai22}
}
\end{document}

%% file: approach_new.tex
%\textbf{Generative Random Walk  Deviation loss}. 

% Our goal is to have points of the same class form a compact cluster in latent space, well separated from other classes. 

We start this section by the formulation of our \emph{Generative Random Walk Deviation} loss.  We will show later in this section how it can be integrated with both generative ZSL models to improve unseen class recognition and with state-of-the-art deep-GAN models to encourage novel visual generations. 
We denote the generator as $G(s,z)$  and its corresponding parameters as  $\theta_G$.  As in ~\cite{xian2018feature,Elhoseiny_2018_CVPR,elhoseiny2019creativity,felix2018multi}, the semantic representation $s$ can be concatenated with a random vector $z \in \mathbb{R}^Z$ sampled from a Gaussian distribution $p_z=\mathcal{N} (0,1)$ to generate an image for visual art generation or visual features in the case of zero-shot learning. Hence, $ G ( s_k, z)$ is the  generated image / feature from the semantic description $s_k$ of class $k$ and the noise vector $z$.  We denote  the discriminator as $D$ and its corresponding parameters as $\theta_D$. 
The discriminator is trained with two objectives: (1)   predict real for images from the training images and fake for generated ones. (2)  identify the category of the input image.  The discriminator then has two heads. The first head is for binary real/fake classification; $\{0,1\}$ classifier. The second head is a $K^s$-way classifier over the seen classes. 
We denote the  real/fake probability produced by $D$ for an input image as $D^r(\cdot)$, and the classification score of a seen class $k \in \mathcal{S}$  given the image as $D^{s,k}(\cdot)$. 
%We denote the generator as $G(s,u)$:  $\mathbb{R}^Z \times \mathbb{R}^T \xrightarrow{\theta_G} \mathbb{R}^X$ and  the discriminator as $D$ : $\mathbb{R}^X \xrightarrow{\theta_D} \{0,1\} \times\{1 \cdots  K^s\} $
%: $\mathbb{R}^X \xrightarrow{\theta_D} \{0,1\} \times\{1 \cdots  K^s\} $, where   and $\theta_D$ are parameters of the generator and the discriminator as  respectively,  $\{ 1 \cdots K^s\}$ is the set of seen class/style labels; see Fig.~\ref{fig1approach}.

\vspace{-1mm}
\subsection{Generative Random Walk Deviation   Loss}
% \kai{we're trying to improve the performance of tf-vaegan on attribute-based datasets.}

 We sample $N_u$ examples that we aim them to deviate from the seen classes/styles in the current minibatch with the generator $G(\cdot)$  . We denote the features of the these hallucinated generations as $X_u = \{x^u_1 \cdots x^u_{N_u}\} $. These features are extracted by   $\phi(\cdot)$, a feature extraction function that we define as the activations from the last layer of the Discriminator $D$ followed by scaled L2 normalization $L2(\mathbf{v},\beta) = \beta \frac{ \mathbf{v}}{\| \mathbf{v} \|}$. The scaled factor  is mainly to amplify the norm of the vectors to avoid the vanishing gradient problem inspired from~\cite{bell2016inside}. We used $\beta=3$ guided by~\cite{bell2016inside,zhang2019large}. 
% We represent seen classes in the same feature apace as $X_u$, as $C = \{ \mathbf{c}_1 \cdots \mathbf{c}_{K^s}\}$, where $\mathbf{c}_i \in \mathbb{R}^{D}$ represents center of class $i$ that we defined as $\mathbf{c}=\phi(G(z = \mathbf{0}, s_i))$,  where $s_i$ is the attribute or text description of seen class $i$. 
We denote the seen class centers that we aim to deviate from as $C = \{ \mathbf{c}_1 \cdots \mathbf{c}_{K^s}\}$, defined in the same feature space as $X_u$, where $\mathbf{c}_i$ represents center of seen class/style $i$. The formulation of $\mathbf{c}_i$ depends on the application (e.g., zero-shot learning or novel art generation), defined later in this section.  
%We represent seen classes as $C = \{ \mathbf{c}_1 \cdots \mathbf{c}_{K^s}\}$, where $\mathbf{c}_i$ represents center of class/style $i$ that we defined as
%\begin{equation}
%    \mathbf{c}_i=\phi(G(z = \mathbf{0}, s_i))
%\end{equation}
% where $s_i$ is the attribute or text description of seen class $i$. 

Let $B \in \mathbb{R}^{N_u\times K^s}$ be the similarity matrix between each of the features of the generations ($x^u \in X_u$ ) and the cluster centers ($c \in C$ ). Similarly, let $A \in \mathbb{R}^{N_u \times N_u }$ compute the similarity matrix between the generated points. In particular,  we use the negative Euclidean distances between the embeddings as a similarity measure as follows:  
\begin{equation}
    B_{ij} = -\|x_i - \mathbf{c}_j \|^2, \,\,\, A_{i,j} = -\|x^u_i - x^u_j\|^2
\end{equation}
where $x^u_i$ and $x^u_j$ are $i^{th}$ and  $j^{th}$ features in the set $X_u$; see Fig.~\ref{fig1approach}. To avoid self-cycle, The diagonal entries $A_{i,i}$ are set to a small number $\epsilon$.  
%Based on A and B, we defined three transition probability matrices $P^{\mathbf{C} \rightarrow X_u} = \textup{softmax}(B^\mathbb{T})$. The softmax operator is applied over each row of the input $B^T$ matrix and hence  $P^{\mathbf{C} \rightarrow X_u}$  represents the transition from each seen class over the $N_u$ generated points. We similarly define $P^{ X_u \rightarrow \mathbf{C }} =  softmax (B)$ as the transition probability from each point over the $K_s$ seen classes. The last transition probability matrix is  $P^{ X_u \rightarrow X_u} = softmax(A)$ from each generated point to over other generated points while avoiding self-loops as we mentioned earlier.  We define our generative random walker probability matrix as: 
Hence, 
we defined three transition probability matrices:
\begin{equation}
\small
   P^{\mathbf{C} \rightarrow X_u} = {\sigma}(B^\mathbb{T}),\,\,\,  P^{ X_u \rightarrow \mathbf{C }} =  \sigma (B),\,\,\, P^{ X_u \rightarrow X_u} = \sigma(A) 
\end{equation}
where $\sigma$ is the softmax operator is applied over each row of the input matrix,  
\noindent  $P^{\mathbf{C} \rightarrow X_u}$ and $P^{ X_u \rightarrow \mathbf{C }}$ are the transition probability matrices from each seen class over the $N_u$ generated points and vice-versa respectively.  $P^{ X_u \rightarrow X_u}$ is the transition probability matrix from each generated point over other generated points. 
%\noindent  $P^{ X_u \rightarrow \mathbf{C }} =  softmax (B)$:  from each generated point over the $K_s$ seen classes. 
%\noindent  $P^{ X_u \rightarrow X_u} = softmax(A)$ from each generated point to over other generated points.% while avoiding self-loops as we mentioned earlier. 
We hence define our generative random walker probability matrix as: 
\begin{equation}\label{eq:rw}
\small
    P^{\mathbf{C } \rightarrow \mathbf{C }} (t, X_u) = {\sigma}(B^\mathbb{T}) \cdot ( \sigma(A))^t \cdot \sigma (B)  
\end{equation}
where  $P^{\mathbf{C } \rightarrow \mathbf{C }}_{i,j} (t, X_u)$ denotes the probability of ending a random walk of a length $t$ at a seen class $j$ {given} that we have started at seen class $i$;   $t$ denotes the number of steps taken between the generated points, before stepping back to land on a seen class/style. 

% \textbf{Loss: } We construct a random walk loss that encourages unseen visual spaces to deviate from seen classes. Hence we defined our loss by encouraging each row in   $P^{\mathbf{C } \rightarrow \mathbf{C }} (t)$ to be hard to classify to seen classes as follows 

\noindent \textbf{Loss.}  Our random walk loss aims at boosting the deviation of unseen visual spaces from seen classes. Hence, we define our loss by encouraging each row in   $P^{\mathbf{C } \rightarrow \mathbf{C }} (t)$ to be hard to classify to seen classes as follows 
\begin{equation}
\small
\begin{aligned}
  {L}_{GRW}(X_u) =& - \sum_{t=0}^{T} {\gamma^{t} \cdot  \sum_{i=1}^{K^s} \sum_{j=1}^{K^s} {U}_c (j) log( {P_{i,j}^{\mathbf{C } \rightarrow \mathbf{C }}}(t,X_u) )}\\
  &-  \sum_{j=1}^{N_u} {U}_x(j)    log(P_v(j))
%   \textcolor{black}{\mathcal{L}_{GRW}  =} & \mathcal{L}_{GRW} + \mathcal{L}_{visit}, 
  \end{aligned}
\label{eq:RW_loss}
\vspace{-2mm}
\end{equation}
% \vspace{-2mm}
where first term minimizes cross entropy loss
between every row in $P^{\mathbf{C } \rightarrow \mathbf{C }}(t,X_u) \forall t=1 \to T$  and uniform distribution over seen classes ${U}_c(j) = {\frac{1}{K^s}, \forall j=1 \cdots K^s}$, where $T$ is a hyperparameter and  $\gamma$ is exponential decay set to $0.7$ in our experiments.  In the  second term, we maximizes the probability of  all the generations  $x^u_i \in X_u$ to be equality visited by the random walk. 
Note that, if we replaced ${U}_c$ by an identity matrix to encourage landing to the starting seen
class, the loss becomes an attraction signal similar to ~\cite{8099557}, which defines its conceptual difference to GRaWD. \emph{We call this version GRaWT, T for aTraction}. 
The second term is called `visit loss' was proposed in~\cite{8099557} to encourage random walker to visit a large set of unlabeled points.  %We applied it here to our generated deviation examples $X_u$.
We compute the overall probability that each generated point would be visited by any of the seen class  $P_v = \frac{1}{N_u} \sum_{i=0}^{N_c}{P^{\mathbf{C} \rightarrow X_u}_i }$, where $P^{\mathbf{C} \rightarrow X_u}_i$ represents the $i^{th}$ row of the $P^{\mathbf{C} \rightarrow X_u}$ matrix; see Fig.~\ref{fig1approach}. The visit loss is then defined  as the  cross-entropy between $P_v$ and the uniform distribution ${U}_x(j) = {\frac{1}{N^u}, \forall j=1 \cdots N^u}$. Hence, visit loss encourages to visit as many examples as possible from $X_u$ and hence improves learning representation. 

\vspace{0mm}
\subsection{GRaWD Integration with Generative ZSL}  
% \vspace{-1.5mm}
Let's denote the set of seen and unseen class labels as $\mathcal{S}$ and $\mathcal{U}$, where $\mathcal{S} \cap \mathcal{U} = \emptyset$. We denote the semantic representations of unseen classes and seen classes as $s_u=\psi(T_u) \in \mathcal{T}$ and $s_i=\psi(T_i) \in \mathcal{T}$ respectively, where $\mathcal{T}$ is the semantic space and $\psi(\cdot)$ is the semantic description function that extract features from text article or attribute description of class $k$. 
Let's denote the seen data as $D^s = \{(x_i^s, y_i^s, s_i) \}$,  where $x_i^s \in \mathcal{X}$ denotes the visual features of the $i^{th}$ image, $y_i^s \in \mathcal{S}$ is the corresponding seen category label. 
For unseen classes, we are given only their semantic representations, one per class, $s_u$. We define  $K^u$ as the number of unseen classes. In Generalized ZSL (GZSL), we aim to predict the label  $y \in \mathcal{U} \cup \mathcal{S}$ at test time given  $x$ that may belong to seen or unseen classes.  We represent seen classes as $C = \{ \mathbf{c}_1 \cdots \mathbf{c}_{K^s}\}$, where $\mathbf{c}_i$ represents the center of class $i$ that we define as
\begin{equation}
    \mathbf{c}_i=\phi(G(z = \mathbf{0}, s_i))
\end{equation}
 where $s_i$ is the attribute or text description of seen class $i$.
$X_u = \{x^u_1 \cdots x^u_{N_u}\} $ are  sampled by $\phi(G (z, s_u))$ where $z \sim
 p_z=\mathcal{N}(0,I)$,  $s_u  \sim p_u$ is a semantic description of a hallucinated unseen class. We explicitly explore the unseen/creative space of the generator $G$ with a hallucinated semantic representation $s_u\sim p^u$, where $p^u$ is a probability distribution over unseen classes, aimed to be likely hard negatives to seen classes. We sample  $s_u \sim p^u$ following the strategy proposed in~\cite{elhoseiny2019creativity} due to its simplicity and effectiveness. It picks two seen semantic descriptions at random $s_a, s_b \in \mathcal{S}$. We then sample  $s^u= \alpha s_a + (1-\alpha) s_b$, 
where $\alpha$ is uniformly sampled between $0.2$ and $0.8$. Note that of $\alpha$ values near $0$ or $1$ are discarded to avoid sampling semantic descriptions that are very close to seen classes. We then integrate  ${L}_{GRW}(X_u)$ with the \emph{Generator $G$  loss} as follows. %sampled by $\phi(G (z, s_u))$ where $z \sim p_z=\mathcal{N}(0,I)$,  $s_u  \sim p_u$ is a semantic description of a hallucinated unseen class;  $p_u$ is defined in the next section.
% \noindent \textbf{ Generator $G$ loss. }
% \vspace{-0.5em}
\begin{equation}
\small 
\begin{aligned}
L_G =& {\lambda \mathbb{E}_{X_u \sim \phi(G({s_u}, z))), z \sim p_{z}, s_u \sim {p^u}} [{L}_{GRW}(X_u)]}\\
&- \mathbb{E}_{z \sim {p_{z}, s_u \sim {p^u}}}[D^r(G(s_u, z))]    \\  
&- \mathbb{E}_{z \sim {p_{z}, (s_k,y^s) \sim {p^s}}}[D^r(G(s_k, z))\\
&+ \sum_{k=1}^{K^s} y^s_k log(D^{s,k}(G(s_k, z)))]
%&+  \frac{1}{K^s}\sum_{k=1}^{K^s}|| \mathbb{E}_{z \sim {p_{z}}}[G(s_k, z)] - \mathbb{E}_{x \sim {p^k_{data}}}[x]||^2 
\label{eq:gen}
\end{aligned}
% \vspace{-2mm} 
\end{equation}
Here, the first term is the proposed GRaWD loss. 
The second  and the third terms trick the generator into classifying the visual generations from both the seen semantic descriptions $s_k$ and unseen semantic descriptions $s_u$, as real.
The fourth term encourages the generator to discriminatively generate visual features conditioned on a given seen class description.  We then define the \emph{Discriminator $D$  loss} as
% \noindent \textbf{Discriminator $D$ loss. }
\begin{equation}
\small 
\begin{aligned}
L_D = \hspace{0.1cm}  
 \mathbb{E}_{z \sim {p_{z}, s_u \sim {p^u}}}&[D^r(G(s_u, z)) ] \,\,\,\,\,\,\,\,\,\,\,\\
  + \mathbb{E}_{z \sim {p_{z}}, (s_k,y^s) \sim {p^s}}&[D^r(G(s_k, z))] - \mathbb{E}_{x \sim {p_{d}}}[D^r(x)] \\
+  L_{Lip} -  \frac{1}{2}  \mathbb{E}_{x,y \sim {p_{d}}}&[\sum_{k=1}^{K^s} y_k log(D^{s,k}(x)) ] \\
-  \frac{1}{2} \mathbb{E}_{z\sim p_z, (s_k,y^s) \sim {p^s}}&[\sum_{k=1}^{K^s} y^s_k log(D^{s,k}(G(s_k, z))) ] \\ 
\end{aligned}
\label{eq:disc}
% \vspace{-2mm} 
\end{equation}
Here, image $x$ and corresponding class one-hot label $y$ are  sampled from the data distribution $p_d$. 
%where $y$ is a one-hot vector encoding of the seen class label for the sampled image $x$
$s_k$ and $y^s$ are
features of a semantic description  
and the corresponding one-hot label sampled from seen classes $p^s$. 
The first three terms approximate Wasserstein distance of the distribution of real features and fake features, and fourth term is the gradient penalty to enforce the Lipschitz constraint: $L_{Lip} =  (||\bigtriangledown_{\tilde{x}} D^r(\tilde{x})||_2 - 1)^2$, where  $\tilde{x}$ is the linear interpolation of the real feature $x$ and the fake feature $\hat{x}$; see~\cite{Gulrajani2017improved}.  The last two terms are the classification losses of the  real and generated data to their corresponding classes.  

%The last two terms are the classification losses of the  real data and features generated from  the semantic descriptions of the seen classes.%; see supplementary for more  \emph{{Implementation details}}.
% \noindent \textbf{Test Time.}  We use $G$ to hallucinate generations for unseen classes given their semantic description, where NN-classification can be applied. 
% % \kai{To say sth: different from classifier-based models?}   %conventional classifier could be trained or NN-classification can be applied.    
% We denote this setting as \textit{inductive ZSL}.
 \subsection{GRaWD Integration with StyleGANs for Novel Art Generation}
 We integrated our loss with DCGAN~\cite{radford2015}, StyleGAN ~\cite{karras2019style} and {StyleGAN2}~\cite{karras2020analyzing} by simply adding $L_{GRW}$ in Eq.~\ref{eq:RW_loss} to the generator loss.  %We integrated our loss with DCGAN~\cite{radford2015}, StyleGAN ~\cite{karras2019style} and {StyleGAN2}~\cite{karras2020analyzing} by adding \textcolor{red}{the first two terms in Eq.~\ref{eq:gen} to the generator loss and the second term to the discriminator loss.} 
 We assume to have $N_s$ seen art styles that we aim to deviate from. Here, we define  $C = \{ \mathbf{c}_1 \cdots \mathbf{c}_{K^s}\}$ by sampling a small episodic memory of size $m$  for every class and computing $\mathbf{c}_i$ from discriminator features.  We randomly sample $m=10$ examples per class  once and compute its mean representation at each iteration. We provide more training details in supplementary.

%% file: main.bbl
\begin{thebibliography}{74}
\providecommand{\natexlab}[1]{#1}

\bibitem[{Akata et~al.(2016)Akata, Perronnin, Harchaoui, and
  Schmid}]{akata2016label}
Akata, Z.; Perronnin, F.; Harchaoui, Z.; and Schmid, C. 2016.
\newblock Label-embedding for image classification.
\newblock \emph{PAMI}, 38(7): 1425--1438.

\bibitem[{Akata et~al.(2015)Akata, Reed, Walter, Lee, and
  Schiele}]{akata2015evaluation}
Akata, Z.; Reed, S.; Walter, D.; Lee, H.; and Schiele, B. 2015.
\newblock Evaluation of output embeddings for fine-grained image
  classification.
\newblock In \emph{{CVPR}}.

\bibitem[{Ayyad et~al.(2020)Ayyad, Navab, Elhoseiny, and
  Albarqouni}]{ayyad2020semi}
Ayyad, A.; Navab, N.; Elhoseiny, M.; and Albarqouni, S. 2020.
\newblock Semi-Supervised Few-Shot Learning with Prototypical Random Walks.

\bibitem[{Bell et~al.(2016)Bell, Lawrence~Zitnick, Bala, and
  Girshick}]{bell2016inside}
Bell, S.; Lawrence~Zitnick, C.; Bala, K.; and Girshick, R. 2016.
\newblock Inside-outside net: Detecting objects in context with skip pooling
  and recurrent neural networks.
\newblock In \emph{Proceedings of the IEEE conference on computer vision and
  pattern recognition}, 2874--2883.

\bibitem[{Briot, Hadjeres, and Pachet(2017)}]{briot2017deep}
Briot, J.-P.; Hadjeres, G.; and Pachet, F. 2017.
\newblock Deep Learning Techniques for Music Generation-A Survey.
\newblock \emph{arXiv:1709.01620}.

\bibitem[{Changpinyo et~al.(2016)Changpinyo, Chao, Gong, and
  Sha}]{changpinyo2016synthesized}
Changpinyo, S.; Chao, W.-L.; Gong, B.; and Sha, F. 2016.
\newblock Synthesized classifiers for zero-shot learning.
\newblock In \emph{{CVPR}}, 5327--5336.

\bibitem[{Chao et~al.(2016)Chao, Changpinyo, Gong, and Sha}]{chao2016empirical}
Chao, W.-L.; Changpinyo, S.; Gong, B.; and Sha, F. 2016.
\newblock An Empirical Study and Analysis of Generalized Zero-Shot Learning for
  Object Recognition in the Wild.
\newblock In \emph{{ECCV}}, 52--68. Springer.

\bibitem[{Chen et~al.(2021)Chen, Wang, Xia, Peng, You, Zheng, and
  Shao}]{chen2021free}
Chen, S.; Wang, W.; Xia, B.; Peng, Q.; You, X.; Zheng, F.; and Shao, L. 2021.
\newblock FREE: Feature Refinement for Generalized Zero-Shot Learning.
\newblock \emph{arXiv preprint arXiv:2107.13807}.

\bibitem[{Date, Ganesan, and Oates(2017)}]{Date2017Fashioning}
Date, P.; Ganesan, A.; and Oates, T. 2017.
\newblock Fashioning with Networks: Neural Style Transfer to Design Clothes.
\newblock In \emph{KDD ML4Fashion workshop}.

\bibitem[{DiPaola and Gabora(2009)}]{dipaola2009incorporating}
DiPaola, S.; and Gabora, L. 2009.
\newblock Incorporating characteristics of human creativity into an
  evolutionary art algorithm.
\newblock \emph{Genetic Programming and Evolvable Machines}, 10(2): 97--110.

\bibitem[{Dumoulin et~al.(2017)Dumoulin, Shlens, Kudlur, Behboodi, Lemic,
  Wolisz, Molinaro, Hirche, Hayashi, Bagan et~al.}]{dumoulin2016learned}
Dumoulin, V.; Shlens, J.; Kudlur, M.; Behboodi, A.; Lemic, F.; Wolisz, A.;
  Molinaro, M.; Hirche, C.; Hayashi, M.; Bagan, E.; et~al. 2017.
\newblock A learned representation for artistic style.
\newblock \emph{ICLR}.

\bibitem[{Elgammal et~al.(2017{\natexlab{a}})Elgammal, Liu, Elhoseiny, and
  Mazzone}]{can_2017}
Elgammal, A.; Liu, B.; Elhoseiny, M.; and Mazzone, M. 2017{\natexlab{a}}.
\newblock CAN: Creative adversarial networks, generating" art" by learning
  about styles and deviating from style norms.
\newblock \emph{arXiv preprint arXiv:1706.07068}.

\bibitem[{Elgammal et~al.(2017{\natexlab{b}})Elgammal, Liu, Elhoseiny, and
  Mazzone}]{elgammal2017can}
Elgammal, A.; Liu, B.; Elhoseiny, M.; and Mazzone, M. 2017{\natexlab{b}}.
\newblock CAN: Creative adversarial networks, generating" art" by learning
  about styles and deviating from style norms.
\newblock In \emph{International Conference on Computational Creativity}.

\bibitem[{Elhoseiny and Elfeki(2019)}]{elhoseiny2019creativity}
Elhoseiny, M.; and Elfeki, M. 2019.
\newblock Creativity Inspired Zero-Shot Learning.
\newblock In \emph{Proceedings of the IEEE International Conference on Computer
  Vision}, 5784--5793.

\bibitem[{Elhoseiny, Saleh, and Elgammal(2013)}]{elhoseiny2013write}
Elhoseiny, M.; Saleh, B.; and Elgammal, A. 2013.
\newblock Write a classifier: Zero-shot learning using purely textual
  descriptions.
\newblock In \emph{{ICCV}}.

\bibitem[{Elhoseiny et~al.(2017)Elhoseiny, Zhu, Zhang, and
  Elgammal}]{Elhoseiny_2017_CVPR}
Elhoseiny, M.; Zhu, Y.; Zhang, H.; and Elgammal, A. 2017.
\newblock Link the Head to the "Beak": Zero Shot Learning From Noisy Text
  Description at Part Precision.
\newblock In \emph{{CVPR}}.

\bibitem[{Farhadi et~al.(2009)Farhadi, Endres, Hoiem, and
  Forsyth}]{farhadi2009describing}
Farhadi, A.; Endres, I.; Hoiem, D.; and Forsyth, D. 2009.
\newblock Describing objects by their attributes.
\newblock In \emph{CVPR 2009.}, 1778--1785. IEEE.

\bibitem[{Felix et~al.(2018)Felix, Kumar, Reid, and Carneiro}]{felix2018multi}
Felix, R.; Kumar, V.~B.; Reid, I.; and Carneiro, G. 2018.
\newblock Multi-modal cycle-consistent generalized zero-shot learning.
\newblock In \emph{ECCV}, 21--37.

\bibitem[{Frome et~al.(2013)Frome, Corrado, Shlens, Bengio, Dean, Mikolov
  et~al.}]{frome2013devise}
Frome, A.; Corrado, G.~S.; Shlens, J.; Bengio, S.; Dean, J.; Mikolov, T.;
  et~al. 2013.
\newblock Devise: A deep visual-semantic embedding model.
\newblock In \emph{{NIPS}}, 2121--2129.

\bibitem[{Gatys, Ecker, and Bethge(2016)}]{Gatys2016ImageStyleTransfer}
Gatys, L.~A.; Ecker, A.~S.; and Bethge, M. 2016.
\newblock Image Style Transfer Using Convolutional Neural Networks.
\newblock In \emph{CVPR}.

\bibitem[{Goodfellow et~al.(2014)Goodfellow, Pouget-Abadie, Mirza, Xu,
  Warde-Farley, Ozair, Courville, and Bengio}]{goodfellow2014generative}
Goodfellow, I.; Pouget-Abadie, J.; Mirza, M.; Xu, B.; Warde-Farley, D.; Ozair,
  S.; Courville, A.; and Bengio, Y. 2014.
\newblock Generative adversarial nets.
\newblock In \emph{{NIPS}}, 2672--2680.

\bibitem[{Gulrajani et~al.(2017)Gulrajani, Ahmed, Arjovsky, Dumoulin, and
  Courville}]{Gulrajani2017improved}
Gulrajani, I.; Ahmed, F.; Arjovsky, M.; Dumoulin, V.; and Courville, A. 2017.
\newblock Improved training of wasserstein gans.
\newblock \emph{arXiv preprint arXiv:1704.00028}.

\bibitem[{Guo et~al.(2017{\natexlab{a}})Guo, Ding, Han, and
  Gao}]{guo2017synthesizing}
Guo, Y.; Ding, G.; Han, J.; and Gao, Y. 2017{\natexlab{a}}.
\newblock Synthesizing Samples for Zero-shot Learning.
\newblock In \emph{{IJCAI}}.

\bibitem[{Guo et~al.(2017{\natexlab{b}})Guo, Ding, Han, and Gao}]{guo2017zero}
Guo, Y.; Ding, G.; Han, J.; and Gao, Y. 2017{\natexlab{b}}.
\newblock Zero-shot Learning with Transferred Samples.
\newblock \emph{IEEE Transactions on Image Processing}.

\bibitem[{Ha and Eck(2018)}]{ha2017neural}
Ha, D.; and Eck, D. 2018.
\newblock A Neural Representation of Sketch Drawings.
\newblock \emph{ICLR}.

\bibitem[{Haeusser, Mordvintsev, and Cremers(2017)}]{8099557}
Haeusser, P.; Mordvintsev, A.; and Cremers, D. 2017.
\newblock Learning by Association — A Versatile Semi-Supervised Training
  Method for Neural Networks.
\newblock In \emph{2017 IEEE Conference on Computer Vision and Pattern
  Recognition (CVPR)}, 626--635.

\bibitem[{Han et~al.(2021)Han, Fu, Chen, and Yang}]{han2021contrastive}
Han, Z.; Fu, Z.; Chen, S.; and Yang, J. 2021.
\newblock Contrastive Embedding for Generalized Zero-Shot Learning.
\newblock In \emph{Proceedings of the IEEE/CVF Conference on Computer Vision
  and Pattern Recognition}, 2371--2381.

\bibitem[{Hertzmann(2018)}]{hertzmann2018can}
Hertzmann, A. 2018.
\newblock Can computers create art?
\newblock In \emph{Arts}, volume~7, 18. Multidisciplinary Digital Publishing
  Institute.

\bibitem[{Isola et~al.(2017)Isola, Zhu, Zhou, and
  Efros}]{Isola2016ImageToImage}
Isola, P.; Zhu, J.; Zhou, T.; and Efros, A.~A. 2017.
\newblock Image-to-Image Translation with Conditional Adversarial Networks.
\newblock \emph{CVPR}.

\bibitem[{Johnson, Alahi, and Li(2016)}]{johnson2016perceptual}
Johnson, J.; Alahi, A.; and Li, F. 2016.
\newblock Perceptual Losses for Real-Time Style Transfer and Super-Resolution.
\newblock \emph{ECCV}.

\bibitem[{Karras et~al.(2018)Karras, Aila, Laine, and
  Lehtinen}]{karras2017progressive}
Karras, T.; Aila, T.; Laine, S.; and Lehtinen, J. 2018.
\newblock Progressive Growing of {GAN}s for Improved Quality, Stability, and
  Variation.
\newblock \emph{ICLR}.

\bibitem[{Karras, Laine, and Aila(2019{\natexlab{a}})}]{karras2019style}
Karras, T.; Laine, S.; and Aila, T. 2019{\natexlab{a}}.
\newblock A style-based generator architecture for generative adversarial
  networks.
\newblock In \emph{Proceedings of the IEEE Conference on Computer Vision and
  Pattern Recognition}, 4401--4410.

\bibitem[{Karras, Laine, and Aila(2019{\natexlab{b}})}]{Karras_2019}
Karras, T.; Laine, S.; and Aila, T. 2019{\natexlab{b}}.
\newblock A Style-Based Generator Architecture for Generative Adversarial
  Networks.
\newblock \emph{2019 IEEE/CVF Conference on Computer Vision and Pattern
  Recognition (CVPR)}.

\bibitem[{Karras et~al.(2020)Karras, Laine, Aittala, Hellsten, Lehtinen, and
  Aila}]{karras2020analyzing}
Karras, T.; Laine, S.; Aittala, M.; Hellsten, J.; Lehtinen, J.; and Aila, T.
  2020.
\newblock Analyzing and improving the image quality of stylegan.
\newblock In \emph{Proceedings of the IEEE/CVF Conference on Computer Vision
  and Pattern Recognition}, 8110--8119.

\bibitem[{Kingma and Welling(2013)}]{kingma2013auto}
Kingma, D.~P.; and Welling, M. 2013.
\newblock Auto-encoding variational bayes.
\newblock \emph{arXiv preprint arXiv:1312.6114}.

\bibitem[{Kodirov, Xiang, and Gong(2017)}]{kodirov2017semantic}
Kodirov, E.; Xiang, T.; and Gong, S. 2017.
\newblock Semantic autoencoder for zero-shot learning.
\newblock \emph{arXiv preprint arXiv:1704.08345}.

\bibitem[{Kumar~Verma et~al.(2018)Kumar~Verma, Arora, Mishra, and
  Rai}]{kumar2018generalized}
Kumar~Verma, V.; Arora, G.; Mishra, A.; and Rai, P. 2018.
\newblock Generalized zero-shot learning via synthesized examples.
\newblock In \emph{CVPR}.

\bibitem[{Lampert, Nickisch, and Harmeling(2009{\natexlab{a}})}]{lampert2009}
Lampert, C.~H.; Nickisch, H.; and Harmeling, S. 2009{\natexlab{a}}.
\newblock Learning to detect unseen object classes by between-class attribute
  transfer.
\newblock In \emph{{CVPR}}, 951--958. IEEE.

\bibitem[{Lampert, Nickisch, and
  Harmeling(2009{\natexlab{b}})}]{lampert2009learning}
Lampert, C.~H.; Nickisch, H.; and Harmeling, S. 2009{\natexlab{b}}.
\newblock Learning to detect unseen object classes by between-class attribute
  transfer.
\newblock In \emph{2009 IEEE Conference on Computer Vision and Pattern
  Recognition}, 951--958. IEEE.

\bibitem[{Lampert, Nickisch, and Harmeling(2013)}]{lampert2013attribute}
Lampert, C.~H.; Nickisch, H.; and Harmeling, S. 2013.
\newblock Attribute-based classification for zero-shot visual object
  categorization.
\newblock \emph{IEEE transactions on pattern analysis and machine
  intelligence}, 36(3): 453--465.

\bibitem[{Li et~al.(2019)Li, Sun, Liu, Zhou, Zheng, Chua, and
  Schiele}]{li2019learning}
Li, X.; Sun, Q.; Liu, Y.; Zhou, Q.; Zheng, S.; Chua, T.-S.; and Schiele, B.
  2019.
\newblock Learning to self-train for semi-supervised few-shot classification.
\newblock In \emph{Advances in Neural Information Processing Systems},
  10276--10286.

\bibitem[{Liu et~al.(2020)Liu, Chen, Pan, Ngo, Chua, and
  Jiang}]{liu2020hyperbolic}
Liu, S.; Chen, J.; Pan, L.; Ngo, C.-W.; Chua, T.-S.; and Jiang, Y.-G. 2020.
\newblock Hyperbolic visual embedding learning for zero-shot recognition.
\newblock In \emph{Proceedings of the IEEE/CVF Conference on Computer Vision
  and Pattern Recognition}, 9273--9281.

\bibitem[{Long et~al.(2017)Long, Liu, Shao, Shen, Ding, and Han}]{long2017zero}
Long, Y.; Liu, L.; Shao, L.; Shen, F.; Ding, G.; and Han, J. 2017.
\newblock From Zero-shot Learning to Conventional Supervised Classification:
  Unseen Visual Data Synthesis.
\newblock In \emph{{CVPR}}.

\bibitem[{Machado and Cardoso(2000)}]{machado2000nevar}
Machado, P.; and Cardoso, A. 2000.
\newblock NEvAr--the assessment of an evolutionary art tool.
\newblock In \emph{Proc. of the AISB00 Symposium on Creative \& Cultural
  Aspects and Applications of AI \& Cognitive Science}, volume 456.

\bibitem[{Mirza and Osindero(2014)}]{mirza2014conditional}
Mirza, M.; and Osindero, S. 2014.
\newblock Conditional generative adversarial nets.
\newblock \emph{arXiv preprint arXiv:1411.1784}.

\bibitem[{Mordvintsev, Olah, and Tyka(2015)}]{mordvintsev2015inceptionism}
Mordvintsev, A.; Olah, C.; and Tyka, M. 2015.
\newblock Inceptionism: Going deeper into neural networks.
\newblock \emph{Google Research Blog. Retrieved June}.

\bibitem[{Narayan et~al.(2020)Narayan, Gupta, Khan, Snoek, and
  Shao}]{narayan2020latent}
Narayan, S.; Gupta, A.; Khan, F.~S.; Snoek, C.~G.; and Shao, L. 2020.
\newblock Latent embedding feedback and discriminative features for zero-shot
  classification.
\newblock In \emph{Computer Vision--ECCV 2020: 16th European Conference,
  Glasgow, UK, August 23--28, 2020, Proceedings, Part XXII 16}, 479--495.
  Springer.

\bibitem[{Nobari, Rashad, and Ahmed(2021)}]{nobari2021creativegan}
Nobari, A.~H.; Rashad, M.~F.; and Ahmed, F. 2021.
\newblock Creativegan: editing generative adversarial networks for creative
  design synthesis.
\newblock \emph{arXiv preprint arXiv:2103.06242}.

\bibitem[{Odena, Olah, and Shlens(2017)}]{odena2016conditional}
Odena, A.; Olah, C.; and Shlens, J. 2017.
\newblock Conditional image synthesis with auxiliary classifier gans.
\newblock In \emph{ICML}.

\bibitem[{Packard(1975)}]{packard1975aesthetics}
Packard, S. 1975.
\newblock Aesthetics and Psychobiology by DE Berlyne.
\newblock \emph{Leonardo}, 8(3): 258--259.

\bibitem[{Patterson and Hays(2012)}]{patterson2012sun}
Patterson, G.; and Hays, J. 2012.
\newblock Sun attribute database: Discovering, annotating, and recognizing
  scene attributes.
\newblock In \emph{Computer Vision and Pattern Recognition (CVPR), 2012 IEEE
  Conference on}, 2751--2758. IEEE.

\bibitem[{Qiao et~al.(2016)Qiao, Liu, Shen, and v.~d. Hengel}]{Qiao2016}
Qiao, R.; Liu, L.; Shen, C.; and v.~d. Hengel, A. 2016.
\newblock Less is More: Zero-Shot Learning from Online Textual Documents with
  Noise Suppression.
\newblock In \emph{{CVPR}}.

\bibitem[{Radford, Metz, and Chintala(2015)}]{radford2015}
Radford, A.; Metz, L.; and Chintala, S. 2015.
\newblock Unsupervised representation learning with deep convolutional
  generative adversarial networks.
\newblock \emph{arXiv preprint arXiv:1511.06434}.

\bibitem[{Radford, Metz, and Chintala(2016)}]{radford2015unsupervised}
Radford, A.; Metz, L.; and Chintala, S. 2016.
\newblock Unsupervised representation learning with deep convolutional
  generative adversarial networks.
\newblock \emph{ICLR}.

\bibitem[{Reed et~al.(2016)Reed, Akata, Mohan, Tenka, Schiele, and
  Lee}]{Reeds2016whatwhere}
Reed, S.~E.; Akata, Z.; Mohan, S.; Tenka, S.; Schiele, B.; and Lee, H. 2016.
\newblock Learning What and Where to Draw.
\newblock In \emph{NIPS}.

\bibitem[{Ren et~al.(2018)Ren, Ravi, Triantafillou, Snell, Swersky, Tenenbaum,
  Larochelle, and Zemel}]{ren2018metalearning}
Ren, M.; Ravi, S.; Triantafillou, E.; Snell, J.; Swersky, K.; Tenenbaum, J.~B.;
  Larochelle, H.; and Zemel, R.~S. 2018.
\newblock Meta-Learning for Semi-Supervised Few-Shot Classification.
\newblock In \emph{International Conference on Learning Representations}.

\bibitem[{Sbai et~al.(2018)Sbai, Elhoseiny, Bordes, LeCun, and
  Couprie}]{sbai2018design}
Sbai, O.; Elhoseiny, M.; Bordes, A.; LeCun, Y.; and Couprie, C. 2018.
\newblock DeSIGN: Design Inspiration from Generative Networks.
\newblock In \emph{ECCV workshop}.

\bibitem[{Schonfeld et~al.(2019)Schonfeld, Ebrahimi, Sinha, Darrell, and
  Akata}]{schonfeld2019generalized}
Schonfeld, E.; Ebrahimi, S.; Sinha, S.; Darrell, T.; and Akata, Z. 2019.
\newblock Generalized zero-and few-shot learning via aligned variational
  autoencoders.
\newblock In \emph{Proceedings of the IEEE/CVF Conference on Computer Vision
  and Pattern Recognition}, 8247--8255.

\bibitem[{Skorokhodov and Elhoseiny(2021)}]{skorokhodov2021class}
Skorokhodov, I.; and Elhoseiny, M. 2021.
\newblock Class Normalization for (Continual)? Generalized Zero-Shot Learning.
\newblock In \emph{International Conference on Learning Representations}.

\bibitem[{Tendulkar et~al.(2019)Tendulkar, Krishna, Selvaraju, and
  Parikh}]{tendulkar2019trick}
Tendulkar, P.; Krishna, K.; Selvaraju, R.~R.; and Parikh, D. 2019.
\newblock Trick or TReAT: Thematic Reinforcement for Artistic Typography.
\newblock In \emph{ICCC}.

\bibitem[{Van~Horn et~al.(2015)Van~Horn, Branson, Farrell, Haber, Barry,
  Ipeirotis, Perona, and Belongie}]{Horn2015}
Van~Horn, G.; Branson, S.; Farrell, R.; Haber, S.; Barry, J.; Ipeirotis, P.;
  Perona, P.; and Belongie, S. 2015.
\newblock Building a Bird Recognition App and Large Scale Dataset With Citizen
  Scientists: The Fine Print in Fine-Grained Dataset Collection.
\newblock In \emph{{CVPR}}.

\bibitem[{Vyas, Venkateswara, and Panchanathan(2020)}]{vyas2020leveraging}
Vyas, M.; Venkateswara, H.; and Panchanathan, S. 2020.
\newblock Leveraging seen and unseen semantic relationships for generative
  zero-shot learning.
\newblock In \emph{European Conference on Computer Vision}, 70--86. Springer.

\bibitem[{Wah et~al.(2011)Wah, Branson, Welinder, Perona, and
  Belongie}]{WahCUB_200_2011}
Wah, C.; Branson, S.; Welinder, P.; Perona, P.; and Belongie, S. 2011.
\newblock {The Caltech-UCSD Birds-200-2011 Dataset}.
\newblock Technical Report CNS-TR-2011-001, California Institute of Technology.

\bibitem[{WikiArt(2015)}]{wikiart20}
WikiArt, O. 2015.
\newblock WikiArt Dataset.
\newblock \url{https://www.wikiart.org/}.
\newblock Accessed: 2020-05-30.

\bibitem[{Wu et~al.(2021)Wu, Zhu, Yong, Wei, Jiang, Zhou, and
  Zhou}]{wu2021clothgan}
Wu, Q.; Zhu, B.; Yong, B.; Wei, Y.; Jiang, X.; Zhou, R.; and Zhou, Q. 2021.
\newblock ClothGAN: generation of fashionable Dunhuang clothes using generative
  adversarial networks.
\newblock \emph{Connection Science}, 33(2): 341--358.

\bibitem[{Wundt(1874)}]{wundt1874grundzuge}
Wundt, W.~M. 1874.
\newblock \emph{Grundz{\"u}ge der physiologischen Psychologie}, volume~1.
\newblock W. Engelman.

\bibitem[{Xian et~al.(2016)Xian, Akata, Sharma, Nguyen, Hein, and
  Schiele}]{xian2016latent}
Xian, Y.; Akata, Z.; Sharma, G.; Nguyen, Q.; Hein, M.; and Schiele, B. 2016.
\newblock Latent embeddings for zero-shot classification.
\newblock In \emph{CVPR}, 69--77.

\bibitem[{Xian et~al.(2018{\natexlab{a}})Xian, Lampert, Schiele, and
  Akata}]{gbu}
Xian, Y.; Lampert, C.~H.; Schiele, B.; and Akata, Z. 2018{\natexlab{a}}.
\newblock Zero-shot learning-a comprehensive evaluation of the good, the bad
  and the ugly.
\newblock \emph{PAMI}.

\bibitem[{Xian et~al.(2018{\natexlab{b}})Xian, Lorenz, Schiele, and
  Akata}]{xian2018feature}
Xian, Y.; Lorenz, T.; Schiele, B.; and Akata, Z. 2018{\natexlab{b}}.
\newblock Feature generating networks for zero-shot learning.
\newblock In \emph{CVPR}.

\bibitem[{Zhang et~al.(2017)Zhang, Xu, Li, Zhang, Wang, Huang, and
  Metaxas}]{zhang2016stackgan}
Zhang, H.; Xu, T.; Li, H.; Zhang, S.; Wang, X.; Huang, X.; and Metaxas, D.
  2017.
\newblock StackGAN: Text to Photo-realistic Image Synthesis with Stacked
  Generative Adversarial Networks.
\newblock In \emph{{ICCV}}.

\bibitem[{Zhang et~al.(2019)Zhang, Kalantidis, Rohrbach, Paluri, Elgammal, and
  Elhoseiny}]{zhang2019large}
Zhang, J.; Kalantidis, Y.; Rohrbach, M.; Paluri, M.; Elgammal, A.; and
  Elhoseiny, M. 2019.
\newblock Large-scale visual relationship understanding.
\newblock In \emph{Proceedings of the AAAI Conference on Artificial
  Intelligence}, volume~33, 9185--9194.

\bibitem[{Zhang, Xiang, and Gong(2016)}]{zhang2016learning}
Zhang, L.; Xiang, T.; and Gong, S. 2016.
\newblock Learning a Deep Embedding Model for Zero-Shot Learning.
\newblock In \emph{{CVPR}}.

\bibitem[{Zhang et~al.(2018)Zhang, Che, Ghahramani, Bengio, and
  Song}]{zhang2018metagan}
Zhang, R.; Che, T.; Ghahramani, Z.; Bengio, Y.; and Song, Y. 2018.
\newblock MetaGAN: An Adversarial Approach to Few-Shot Learning.
\newblock In \emph{Advances in Neural Information Processing Systems},
  2371--2380.

\bibitem[{Zhu et~al.(2018)Zhu, Elhoseiny, Liu, Peng, and
  Elgammal}]{Elhoseiny_2018_CVPR}
Zhu, Y.; Elhoseiny, M.; Liu, B.; Peng, X.; and Elgammal, A. 2018.
\newblock A generative adversarial approach for zero-shot learning from noisy
  texts.
\newblock In \emph{CVPR}.

\end{thebibliography}
